\documentclass[11pt,draftcls,onecolumn]{IEEEtran}
\usepackage{amssymb}
\usepackage{multirow,setspace,verbatim,amsfonts,graphicx,amsmath,amsthm,amsbsy,amssymb,epsfig,url}
\usepackage{soul}
\usepackage{amsthm}

\newcommand{\Ab}{{\mathbf A}}

\newcommand{\Eb}{{\mathbf E}}

\newcommand{\Hb}{{\mathbf H}}

\newcommand{\Lb}{{\mathbf L}}
\newcommand{\Mb}{{\mathbf M}}

\newcommand{\Sb}{{\mathbf S}}

\newcommand{\Ub}{{\mathbf U}}
\newcommand{\Vb}{{\mathbf V}}

\newcommand{\Xb}{{\mathbf X}}

\newcommand{\Zb}{{\mathbf Z}}

\newcommand{\mb}{{\mathbf m}}
\newcommand{\nb}{{\mathbf n}}

\newcommand{\rb}{{\mathbf r}}
\renewcommand{\sb}{{\mathbf s}}

\newcommand{\xb}{{\mathbf x}}
\newcommand{\yb}{{\mathbf y}}

\newcommand{\Hc}{\mathcal{H}}

\newcommand{\Thetab}{{\boldsymbol {\Theta}}}
\newcommand{\Lambdab}{{\boldsymbol {\Lambda}}}

\newcommand{\Rd}{{\mathbb R}}

\newcommand{\Ybc}{{\boldsymbol{\mathcal Y}}}

\newcommand{\Zd}{\mathbb{Z}}
\newcommand{\rank}{\textsc{Rank}}
\newcommand{\hank}{\mathscr{H}}
\newcommand{\hnk}{\mathfrak{h}}

\newcommand{\omegab}{{\boldsymbol {\omega}}}

\newcommand{\Sc}{\mathcal{S}}
\newcommand{\zerob}{\mathbf{0}}
\newcommand{\Null}{\textsc{Nul}}

\newcommand{\imsizes}{4.5}%

\usepackage{cite}
\usepackage{makecell}

\ifCLASSINFOpdf

\else

\fi
\usepackage{amsfonts}
\usepackage{amsmath,bm}
\usepackage{amssymb}
\usepackage{amsthm}
\usepackage{tikz,graphicx}
\usepackage{mathrsfs} 
\usepackage{algpseudocode}
\usepackage{algorithm}
\usepackage{array,multirow}
\usepackage{color}
\usepackage{slashbox} 			%%%
\usepackage{diagbox}
\newcommand{\MONTH}{%
  \ifcase\the\month
  \or JANUARY% 1
  \or FEBRUARY% 2
  \or MARCH% 3
  \or APRIL% 4
  \or MAY% 5
  \or JUNE% 6
  \or JULY% 7
  \or AUGUST% 8
  \or SEPTEMBER% 9
  \or OCTOBER% 10
  \or NOVEMBER% 11
  \or DECEMBER% 12
  \fi}

\begin{document}
%\title{Low Rank Reproducing (LRR) Filter for Interpolation of Irregularly sampled Images}
\title{Sparse + Low Rank Decomposition of  Annihilating Filter-based Hankel Matrix for Impulse Noise Removal}

\author{Kyong Hwan Jin and Jong Chul Ye,~\IEEEmembership{Senior Member,~IEEE}% <-this % stops a space
}

%\markboth{IEEE TRANSACTIONS ON IMAGE PROCESSING,~Vol.~, No.~, October~2014}%
%\markboth{IEEE TRANSACTIONS ON IMAGE PROCESSING,~Vol.~, No.~, \MONTH~\the\year}
%\markboth{SUBMITTED TO IEEE TRANSACTIONS ON IMAGE PROCESSING}
%{Shell \MakeLowercase{\textit{et al.}}: Bare Demo of IEEEtran.cls for Journals}

\maketitle

\begin{abstract}
\baselineskip 0.17in
Recently,  so called annihilating filer-based low rank Hankel matrix  (ALOHA) approach was proposed as a powerful image inpainting method.  Based on the observation that  smoothness or textures within an image patch corresponds to  sparse
spectral components in the frequency domain, ALOHA exploits the existence of annihilating filters  and  the associated rank-deficient Hankel matrices in the image domain to estimate the missing pixels.
By extending this idea, here we   propose  a novel impulse noise removal algorithm using sparse + low rank decomposition of  an  annihilating filter-based  Hankel matrix.
The new approach, what we call the robust ALOHA, is  motivated by
%to remove  impulse noises such as  random valued impulse noises and salt/pepper noises.
%Based on 
the observation that an image corrupted with impulse noises  has intact pixels; so the impulse noises can be modeled as sparse components, whereas the underlying image can be still modeled using a low-rank Hankel structured matrix.
To solve the sparse + low rank decomposition problem, we  propose an alternating direction method of multiplier (ADMM) method with initial factorized matrices coming from low rank matrix fitting (LMaFit) algorithm. 
To adapt the local image statistics that have distinct spectral distributions, the robust ALOHA is applied patch by patch. Experimental results from two types of impulse noises - random valued impulse noises and salt/pepper noises - for both single channel and multi-channel color images demonstrate that the robust ALOHA outperforms  the existing algorithms up to 8dB in terms of the peak signal to noise ratio (PSNR).
\end{abstract}


\noindent{
\hspace{-0.2cm}\noindent {\bf Correspondence to:}\\
\vspace{-0.2cm}Jong Chul Ye,  Ph.D. ~~Professor \\\
\vspace{-0.2cm}Dept. of Bio and Brain Engineering,  KAIST \\
\vspace{-0.2cm}291 Daehak-ro Yuseong-gu, Daejon 305-701, Republic of Korea \\
\vspace{-0.2cm}Email: jong.ye@kaist.ac.kr, Tel: 82-42-350-4320  }

\IEEEpeerreviewmaketitle

\newpage
\section{Introduction}

Impulse noises occur by malfunctioning of detector pixels in camera or from the missing memory elements  in imaging hardware \cite{chan2005salt}. There are two types of impulse noises.  The first type includes   salt/pepper noises which have the extremal values of dynamic ranges; therefore, the noise pixels can be easily detected by an adaptive median filter (AMF)\cite{hwang1995adaptive}. The second type noises are random valued impulse noises (RVIN) which have random values within the dynamic range of an image pixel. Unlike the salt/pepper noises, the noisy pixels of RVIN cannot be effectively detected by an adaptive median filter.  Instead, the adaptive center-weighted median filter (ACWMF)\cite{chen2001adaptive}  has been widely used to find the locations of noisy pixels. Even with AMF and ACWMF, when the density of noise increases, the denoising performance of these single step algorithms  become severely degraded. To compensate for this weakness, two-phase denoising algorithms with ``decision-based filter'' or ``switching filter''  were proposed \cite{yan2013restoration,chen2001space,eng2001noise,pok2003selective,chan2005salt}. 
More specifically, these algorithms consist of two main parts: detecting noise pixels by AMF, ACWMF, or other outlier finding algorithms; and then replacing the detected noise pixels with the estimated values using the total variation\cite{yan2013restoration} or edge preserving regularizations \cite{chan2005salt,nikolova2004variational}, while leaving other noiseless pixels unchanged.

On the other hand,  impulse noise denoising algorithms using  proximal optimizations with non-smooth penalties were recently proposed \cite{chambolle2011first,yang2009efficient,chan2011augmented}. In particular,  in the TVL1 (total variation $l_1$) approach \cite{yang2009efficient,chambolle2011first},  the data  fidelity term was measured with the $l_1$ norm  to deal with impulse outliers and the total variation regularization was used as a penalty. They can effectively remove the impulse noises at  sufficiently fast speed.  However, the algorithm  often causes edge distortions  or   texture patterns blurrings due to the TV terms.  With the advance of compressed sensing theory \cite{donoho2006compressed,candes2006robust},  the impulse noise denoising methods based on compressed sensing were also proposed \cite{carrillo2010robust,Majumdar2016136}. In \cite{Majumdar2016136}, the authors encouraged the spatio-spectral domain redundancy to be sparsely represented in Fourier domain by using blind compressed sensing framework. This approach demonstrated outstanding recovery performances; however, the algorithm cannot be performed without highly correlated spectral dataset. In \cite{carrillo2010robust}, the sparsity level of target signal was minimized just as in the conventional CS approach using noisy measurements contaminated by impulse noise. However,  the performance was inferior to the two-phase methods \cite{chan2005salt}. While a low-rank matrix completion approach for impulse noise denoising for video sequence was proposed \cite{ji2010robust}, the algorithm only worked for video sequence denoising,  because spatio-temporal redudancy should be exploited in this method.

One of the unique characteristics of impulse noises compared to Gaussian or Poisson noise  is that
an image corrupted with impulse noises  has intact pixels; so the impulse noises can be modeled as sparse components, whereas the underlying intact image still retains the original image characteristics.
In fact, this was the main idea utilised in TVL1 approach  \cite{yang2009efficient}, and  this paper also aims at fully exploiting this observation.
However, 
one of the most important contributions of this paper is to demonstrate that there exists a more natural and powerful image modeling method than the TV approach,
 and that the
resulting impulse noise removal problem becomes a {\em sparse + low rank} decomposition problem.
This is  inspired by our novel image modeling and associated inpainting method using so-called the annihilating filter-based low-rank Hankel matrix (ALOHA) approach  \cite{7127011}.

More specifically, %annihilating filer-based low rank Hankel matrix  (ALOHA) approach was proposed as a powerful image inpainting method.  B
in  \cite{7127011},  we demonstrated 
that the smoothness or textures within an image patch leads to  {\em  sparse}
spectrum in the frequency domain,  so the sampling theory of signals with finite rate of innovations  \cite{vetterli2002sampling,maravic2005sampling} tells us that
there exists annihilating filters that annihilate the pixel values within the corresponding  image patch.
Moreover,  the existence of annihilating filter enables us to   construct a rank-deficient Hankel structured matrix
whose rank is determined by the sparsity in the spectral domain \cite{7127011}.
Thanks to this observation,  an image patch could be modeled using an annihilating filter-based Hankel structured matrix (ALOHA), and 
image inpainting problems was solved by a low rank matrix completion algorithm.
The idea was  extended by our group for compressed sensing MRI \cite{jin2015general,7163879}, 
image deconvolution \cite{min2015Fast},  and  interpolation of scanning microscopy \cite{jin2015patch}.
Ongie {\em et al} independently developed similar  approaches for  super-resolution MRI\cite{ongie2015super,ongie2015recovery}.

One of the most important consequences  of ALOHA in the context of impulse noise removal is an observation saying that the construction of Hankel structured matrix is a kind of  {\em linear} lifting scheme, so the sparse components in image  are also
sparse in the lifted Hankel matrix. 
Therefore, we can use a sparse+low rank decomposition of the Hankel structured matrix to decouple the sparse impulse noise components from the underlying image.
The new algorithm, what we call robust ALOHA, is applied patch by patch to adapt the local image statistics  that have distinct spectral distribution.
We are aware that   there have been significant progresses on the decomposition of superposed matrix consisting of low-rank and sparse components \cite{candes2011robust,wright2009robust,tao2011recovering,lin2010augmented,chartrand2012nonconvex},
which is often called Robust Principal Component Analysis (RPCA) \cite{candes2011robust}. 
However, the matrix in RPCA is usually unstructured, whereas the robust ALOHA uses an Hankel structured matrix.
As will be shown later, the introduction of Hankel structure matrix significantly improves the denoising performance by exploiting the spectral domain sparsity. 

During the writing of this paper, we came across a very interesting sparse and low rank decomposition of Hankel structure matrix for predicting target locations under the occlusion\cite{ayazoglu2012fast}. While the optimization framework and algorithm used in \cite{ayazoglu2012fast} have similarities with ours, the idea of \cite{ayazoglu2012fast} was originally inspired by the dynamic system identification rather than image modeling using annihilating filter and we are not aware of its applications for impulse noise removal. Therefore, we believe that the application of robust ALOHA to impulse noise removal is sufficiently novel.

To solve the associated sparse+low rank decomposition problem of Hankel structure matrix, 
an alternating direction method of multiplier (ADMM)\cite{boyd2011distributed} is utilized with initial factorized matrices from low rank matrix fitting algorithm (LMaFit) \cite{wen2012solving}.
%Moreover, when the locations of noises are available, for example, by AMF or ACWMF for the case of salt/pepper noise, the proposed algorithm can be easily modified to exploit this information.
Furthermore, the denoising algorithm is also extended to exploit the joint sparsity constraint in  colour images  by stacking Hankel structured matrix from each channel side by side and applying sparse+low rank decomposition
of the concatenated Hankel matrix.
 Using extensive numerical experiments,  we demonstrate that
 the robust ALOHA significantly outperforms all the existing methods.
 
This paper is organized as follows. Section \ref{sec:sparselowrank} discusses the theory behind the robust ALOHA.  In Section \ref{sec:sl_hankel}, an optimization method for the associated sparse+low rank decomposition
of Hankel structured matrix is described.  
Extension to multi-channel denoising is discussed in Section~\ref{sec:color}. Experimental results are provided in Section \ref{sec:result}, which is followed by  discussion and conclusion in Section \ref{sec:discussion} and \ref{sec:conclusion}.

\section{Theory}\label{sec:sparselowrank}

\subsection{Review of TVL1 approach}
Because impulse noises occur by malfunctioning of detector or memory elements   \cite{chan2005salt},
 only a subset of image pixels are corrupted by the noises. 
Therefore, if  $\Mb$ denotes an image measurement corrupted by impulse noises, it can be modeled as
$$\Mb = \Xb+\Eb$$
where $\Xb$ is the underlying ``clean" image, and $\Eb$ denotes a sparse matrix composed of impulse noises.
This model is quite often used in the existing impulse noise removal algorithm.  For example, in TVL1 \cite{yang2009efficient,chambolle2011first}, 
$\Eb$ is considered as a sparse outlier,  whereas the underlying image is modeled using the total variations. This leads to the following minimization problem:
\begin{eqnarray}\label{eq:TV}
\|\Mb-\Xb\|_{1} + \lambda TV(\Xb)
\end{eqnarray}
where the $\|\cdot\|_{1}$ norm
is the $l_1$ norm corresponding to summation of absolute values of each matrix elements for outlier removal,
and the $TV(\Xb)$ denotes the 2-D TV penalty to model the underlying image.
In the following, we will explain how the image model in \eqref{eq:TV} is modified in the proposed method to give superior denoising performance.

\subsection{Image modeling using low-rank Hankel structured matrix}

\begin{figure}[!bt]
\centering
\includegraphics[width=12cm]{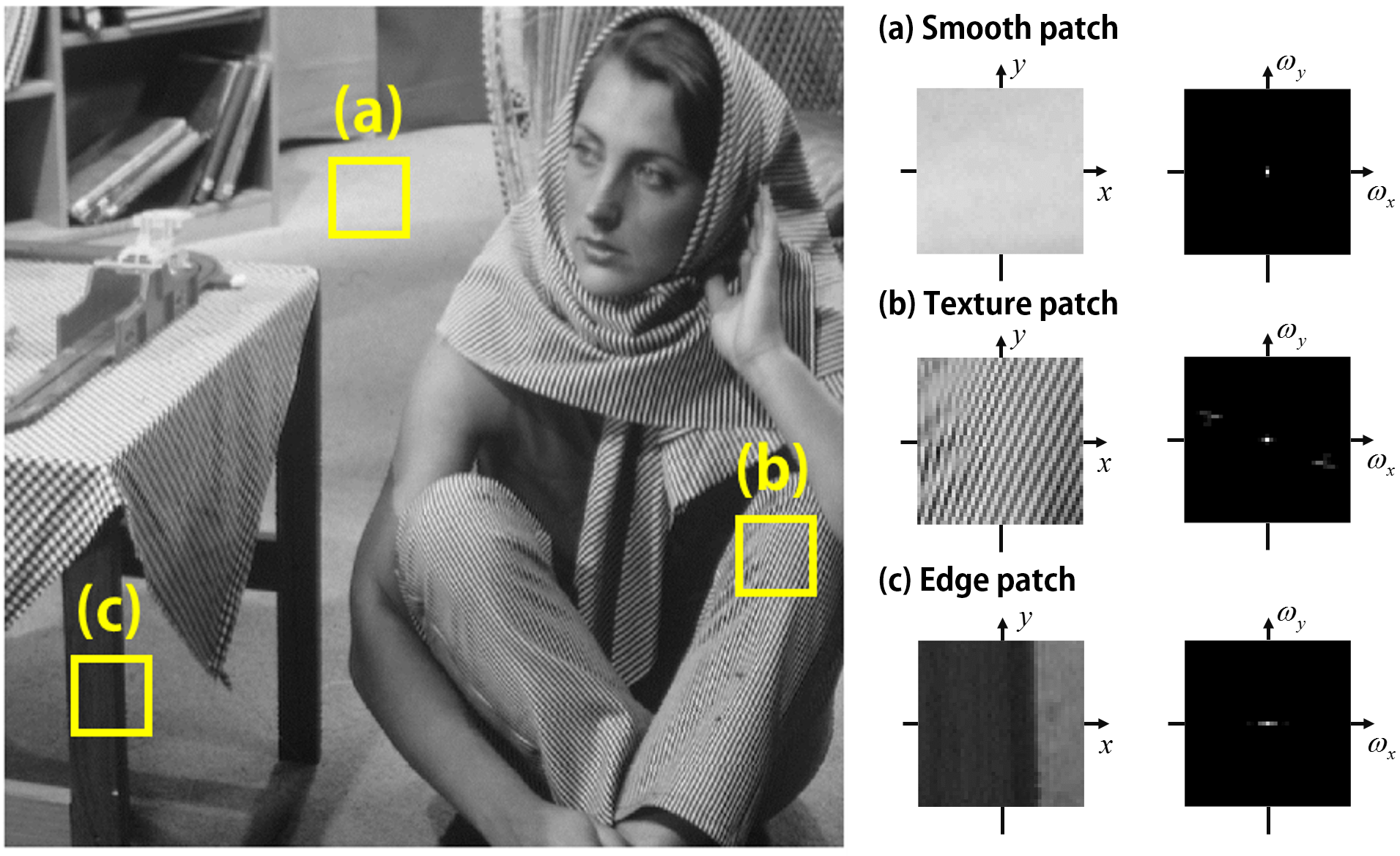}
\caption{Spectral components of patches from  (a) smooth background,  (b) texture, and (c) edge.}%The sampled spectrum of a patch can be represented in (c).}
\label{fig:flowchart}
\end{figure}

In our recent work  \cite{7127011}, we demonstrate that   diffusion \cite{chan2001nontexture,catte1992image,alvarez1992image,black1998robust} and/or   Gaussian Markov random field (GMRF)  approaches for image modelling \cite{cross1983markov,li2008markov,chellappa1985classification,cohen1991classification,manjunath1991unsupervised}  are closely related to an annihilating filter relationship from
the sampling theory of signals with finite rate of innovations (FRI)  \cite{vetterli2002sampling,maravic2005sampling}. 
More specifically,  as shown in Fig. \ref{fig:flowchart}(a),  a smoothly varying patch usually has spectrum content in the low frequency regions, and the other frequency regions have very little spectral components.
A similar spectral domain sparsity can be observed in a texture patch in  Fig. \ref{fig:flowchart}(b), where the spectral components are mainly concentrated on the fundamental frequencies of the patterns.
For the case of an abrupt transition along the edge as shown in Fig. \ref{fig:flowchart}(c), the spectral components are mostly localized along the $\omega_x$ axis.

Mathematically,
when a patch $x(\rb)\in \Rd^2$ has sparse {\em spectral} components, we can show that there exists a corresponding annihilating filter in the {\em image} domain. 
More specifically, if the spectrum of an image patch is described by
\begin{gather}
%\mathcal{F}\{x(\rb)\}=
\hat{x}(\boldsymbol{\omega})=\sum^{k-1}_{j=0}c_j \delta(\boldsymbol{\omega}-\boldsymbol{\omega}_j), ~~\boldsymbol{\omega}=(\omega_x,\omega_y) 
\end{gather}
where $k$ denotes the number of non-zero spectral components, % and ${\cal F}(\cdot)$ denotes the Fourier transform.
then  it is easy to find an annihilating function $\hat h(\omegab)$ in the spectrum domain that does not overlap with the support of  $\hat x(\omegab)$, i.e. 
\begin{equation}\label{eq:annf}
\hat h(\omegab)\hat x(\omegab) = 0  \, \quad \forall \omegab.
\end{equation}
This implies the existence of the annihilating filter $h(\rb) := \mathcal{F}^{-1}\{ \hat h(\omegab) \} $ in the image domain:
\begin{equation}\label{eq:annf2}
 h(\rb)\ast x(\rb) = 0  .
\end{equation}
For example, if an image is sufficiently flat with little variations in pixel values, then its spectral component $\hat f(\omegab)$ is most concentrated around the zero frequency component (i.e. $\hat x(\omegab)\simeq 0, \omegab\neq 0$); therefore,
$\hat h(\omegab) = -\|\omegab\|^2$ becomes an annihilating function in the spectral domain.
The associated annihilating filtering in the image domain is then given by
\begin{equation}\label{eq:diffusion}
h(\rb) \ast  x(\rb) = \Delta x(\rb).
\end{equation}
where $\Delta$ denotes the Laplacian operator, 
which corresponds to the diffusion operation that is widely used in image denoising, inpainting, and so on \cite{chan2001nontexture,catte1992image,alvarez1992image,black1998robust,chambolle2011first,yang2009efficient,chan2011augmented}.
This example clearly shows why the diffusion based approach is closely related to the annihilating filter approach.

%Now, as  shown in Fig.~\ref{fig:kSparse}(b),
If Fourier measurement data $\hat x(\omegab)$ is discretized at an appropriate Nyquist sampling rates, the corresponding
 discrete counterpart  is given by
\begin{eqnarray}\label{eq:zero}
( h\ast  x)[\nb] =  \sum_{\mb}  h[\mb] x[\nb-\mb]  = 0, \quad  \nb,\mb \in \Zd^2 ,
\end{eqnarray}
where the discrete filter $h[\nb]$ is  now a discrete annihilating filter.
Among the various of choices of annihilating function $\hat h(\omegab)$ that satisfies \eqref{eq:annf},
Vetterli et al    \cite{vetterli2002sampling,maravic2005sampling} showed that an annihilating function can be constructed
using a finite combination of sinusoidals such that the corresponding discrete annihilating filter  $h[\nb]$ has a  finite filter length.
In this case, the convolution \eqref{eq:zero} becomes finite length convolution, and we can exploit \eqref{eq:zero} in a matrix representation.

Specifically, let $\Xb \in \Rd^{N\times M}$   denote the matrix composed of  $x[\nb]$ such that
$\Xb_{i,j}= x[i,j]$.  We also define the discrete filter matrix $\Hb \in \Rd^{p\times q}$ such that  $\Hb_{i,j}=h[i,j]$. %We further denote the $i$-th column vector of $\Xb$ as $\xb_i$.
Then,
by removing the boundary data beyond the image patch, we can construct the following matrix equation:
\begin{eqnarray}\label{eq:mateq}
 \hank\{\Xb\} \overline{\textsc{vec}(\Hb)}= \mathbf{0},
\end{eqnarray}
where  $ {\textsc{vec}(\Hb)}$ is the vectorisation of the matrix $\Hb$ and the overline is the order reversal operation.
Moreover, 
the 2-D Hankel matrix $ \hank\{\Xb\}$ in \eqref{eq:mateq} is defined by
\begin{eqnarray}\label{eq:3dhank}
  \hank\{\Xb\} = &    \!\left[\!
        \begin{array}{cccc}
            \! 
        \hnk\{\xb_1\}&   \hnk\{\xb_{2}\} &   \cdots & \hnk\{ \xb_{q}\}\\
         \hnk\{\xb_2\} &   \hnk\{\xb_{3}\} &   \cdots & \hnk\{ \xb_{q+1}\}\\
     %   \0 &    \cdots & \Gb_1^{(k)} & \Gb_2^{(k)} \\
       \vdots &  \vdots  & \ddots  & \vdots \\
 \hnk\{\xb_{N-q+1}\} &    \hnk\{\xb_{N-2n+1}\} & \cdots &\hnk\{\xb_{N}\}  \\
                     \end{array}\!
              \right]\!   ~\in \Rd^{(N-q+1)(M-p+1) \times pq}
\ ,
    \end{eqnarray}
and the 1-D Hankel matrix $\hnk\{\xb_i\}$ for the $i$-th column $\xb_i$  of the matrix $\Xb$ is given by
\begin{equation}\nonumber
\hnk\{\xb_i\}= \left[
        \begin{array}{cccc}
      x[1,i] & x[2,i] & \cdots  & x[p,i] \\
       x[2,i] & x[3,i] & \cdots & x[p+1,i] \\
      \vdots    & \vdots     &  \ddots   &    \vdots   \\
        x[M-p+1,i] & x[M-p+2,i] & \cdots &  x[M,i]\
        \end{array}
    \right] \ .
\end{equation}
If the underlying signal has  $k$-sparse spectral components, we can further show the following key result  \cite{jin2015general}:
\begin{eqnarray}\label{eq:k}
\rank \hank(\Xb)= k, \quad % \quad\leq \min\{n_1-\kappa+1, \kappa\} \ ,
\end{eqnarray}
which implies that as long as the annihilating filter size is bigger than the sparsity level, the rank of the associated Hankel
matrix is always equal to the sparsity level.
This implies the following duality:
\begin{eqnarray}
	\boxed{  \mbox{$k$ spectral sparse image } ~ \overset{\mathcal{F}} \longleftrightarrow ~\mbox{Rank-$k$ 2-D Hankel structure matrix}  } \nonumber%\label{eq:dual}
\end{eqnarray} 
By exploiting this duality, our previous work \cite{7127011}
derived an annihilating filter-based low rank Hankel matrix approach for image inpainting. 
In this paper, this approach will be recasted into sparse+low rank decomposition for impulse noise removal.

\subsection{Sparse + Low-rank Decomposition Model  for Hankel Structured Matrix from Impulse Noises}

\begin{figure}[!bt]
\centering
\includegraphics[width=16cm]{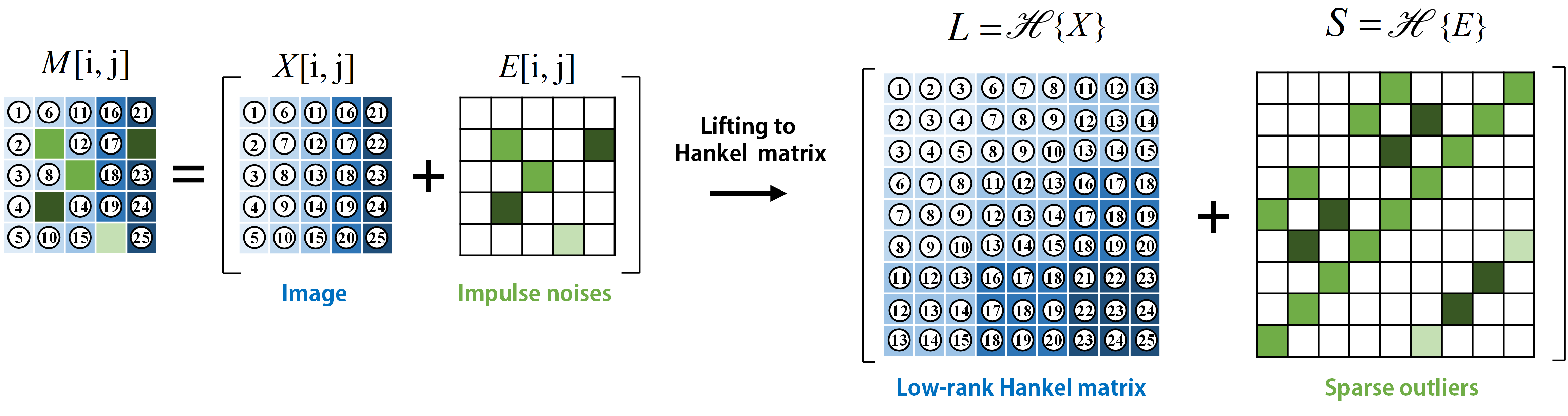}
\caption{Sparse + low rank decomposition of Hankel structured matrix from an image patch corrupted with impulse noise. Because 
a lifting to Hankel structure is linear, the sparse impulse noises are also lifted to  sparse outliers.}%is conducted as from discretized version of Fig. \ref{fig:flowchart} (c).}
\label{fig:flowchart2}
\end{figure}
Unlike the lifting scheme used for phase retrieval  problems \cite{candes2015phase}, our lifting scheme to the Hankel 
structured matrix is linear, so
the { sparse} impulse noise is also lifted to the sparse outliers  in the lifted Hankel structured matrix (see Figure~\ref{fig:flowchart2}).  
Accordingly,  if an underlying image  is corrupted with sparse impulse noises,  
 then we have
\begin{equation}\label{eq:LS}
\hank\{\Xb\} = \Lb + \Sb
\end{equation}
where $\Lb$ denotes the low-rank component and $\Sb$ represents the sparse components originated from impulse noises, which are both
in the Hankel structure.
This is the key property we want to exploit in our impulse noise removal algorithm.
In fact, to address this type of sparse + low rank decomposition,
 the robust principal component analysis (RPCA)  was actively investigated \cite{candes2011robust,wright2009robust,tao2011recovering,lin2010augmented,chartrand2012nonconvex}.
%This means that one can successfully extract principal components from the matrix even though there are partial corrupted values on the matrix. The RPCA can extract principal components without outlier (sparse part) with respect to user defined trade-off variable and minimizes a weighted summation of the nuclear norm and the $l1$ norm\cite{candes2011robust} given as
More specifically, for a given measurement matrix $\Mb$, the RPCA solves the following minimization problem:
\begin{eqnarray}
\min && \|\Lb\|_\ast +\tau\|\Sb\|_1\label{eq:rpca_1}\\
 \mbox{subject to } && \Lb+\Sb=\Mb,\label{eq:rpca_2}
\end{eqnarray}
where $\|\cdot\|_*$ is the nuclear norm. %, $\|\cdot\|_1$ is the $l_1$ norm corresponding to summation of absolute values of input. 
To minimize this, alternating directions methods \cite{yuan2009sparse} were employed. 

\begin{figure}[!hbt]
\centering
\includegraphics[width=12cm,height=8cm]{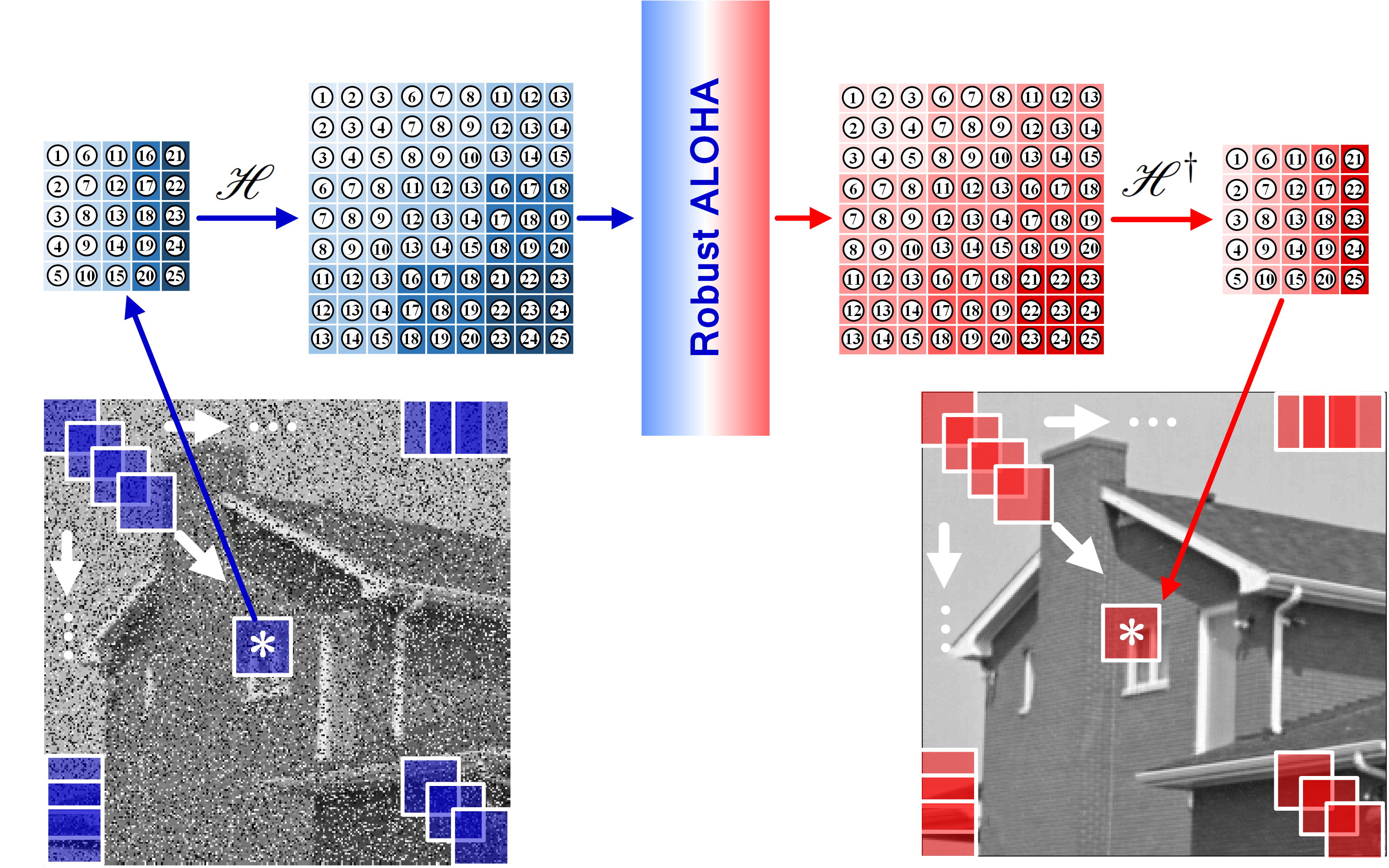}
\caption{Patch-by-patch processing framework using robust ALOHA for impulse noise removal.}
\label{fig:scanning}
\end{figure}

Compared to the standard RPCA approach, our sparse+ low rank decomposition problem using  \eqref{eq:LS} requires additional constraint to impose Hankel structures.
Therefore the RPCA algorithm should be modified.
The specific optimization algorithm under this constraint will be explained in the following sections.
Additionally, because the image statistics change across an image with spatially varying annihilating property,
a noisy image should be partitioned into overlapped  patches, which are processed patch-by-patch using robust ALOHA and the average values are used as described in the algorithm flowchart in Fig. \ref{fig:scanning}.%\cite{7127011}. 
 
\section{Optimization Methods}\label{sec:sl_hankel}

\subsection{Sparse + Low Rank Decomposition for Hankel Matrix}

Note that the Hankel structure matrix in \eqref{eq:3dhank} is determined by the underlying image patch ($\Xb$) size and the associated  annihilating filter  ($\Hb$) size.
For given $M\times N$  image patch and $p\times q$ annihilating filter,  we now denote the associated spaces for the Hankel matrix as 
$\Hc(M,N;p,q)$.
Then,  for a given noisy image patch $\Mb \in \Rd^{M\times N}$ and $p\times q$ annihilating filter size, our impulse noise removal algorithm can be implemented by solving the following
sparse + low rank decomposition under the Hankel structure matrix constraint: % a given measurement matrix $\Mb$, the RPCA solves the following minimization problem:
\begin{eqnarray}\label{eq:robustEMAC}
(P) \quad & \min_{\Lb,\Sb} & \|\Lb\|_\ast +\tau\|\Sb\|_1 \\
& \mbox{subject to } & \Lb+\Sb=\hank\{\Mb\}, \notag\\
 && \Lb, \Sb \in \Hc(M,N;p,q) \notag
\end{eqnarray}
Since the sparse components in image patch are also sparse in a lifted Hankel structure, ($P$) can be further simplified as
%The sparse and low-rank decomposition is started from general equation of RPCA given as
\begin{eqnarray*}
(P')\quad &  \min_{\Lb,\Eb}  & \|\Lb\|_* +\tau\|\Eb\|_1 \nonumber \\
& \mbox{subject to}&  \Lb = \hank\{\Xb\} \\
&&  \Xb +\Eb=\Mb  .\label{eq:data0}
\end{eqnarray*}
where, with a slight abuse of notation,  $\tau$ denotes an appropriately scaled version from $\tau$ in $(P)$. Note that $\Eb$ is now in image patch domain, unlike $\Sb$ in the lifted Hankel matrix structured matrix domain in $(P)$.
The advantage of $(P')$ over $(P)$ is an associated simpler optimization method.
More specifically, if we apply  a factorized form of nuclear norm relaxation \cite{srebro2004learning}, then
 the final problem formulation of the optimization problem  can be expressed as
\begin{eqnarray}
\min_{\Eb,\{(\Ub,\Vb)| \Ub\Vb^H =\hank\{\Xb\}\} }  && \|\Ub\|_F^2+ \|\Vb\|_F^2 +\tau\|\Eb\|_1 \label{eq:fac} \\
\mbox{subject to} %&&\hank\{\Xb\} = \Ub\Vb^H \label{eq:fac}\\
&& \Xb+\Eb=\Mb.\label{eq:data}
\end{eqnarray}
The constraints in \eqref{eq:fac} and \eqref{eq:data} can be handled using alternating direction method of multiplier  (ADMM) \cite{boyd2011distributed,signoretto2013svd,tao2011recovering}. %a set 
The associated Lagrangian function  ADMM is  given by:
\begin{eqnarray}\label{eq:ADMM}
L(\Ub,\Vb,\Eb,\Xb,\Thetab,\Lambdab) :=&& \frac{1}{2} \left( \|\Ub\|_F^2+\|\Vb\|_F^2\right) + \tau \|\Eb\|_1+\frac{\beta}{2} \| \Xb+\Eb-\Mb+\Thetab \|_F^2  + \nonumber\\
 && \frac{\mu}{2}  \|\hank\{\Xb\}- \Ub\Vb^H+\Lambdab\|^2_F
\end{eqnarray}
Then, each subproblem is simply obtained from \eqref{eq:ADMM}. More specifically, we have
 \begin{eqnarray}
 \Eb^{(k+1)}   &=& \arg\min_\Eb \tau \|\Eb\|_1+ \frac{\beta}{2} \| \Xb^{(k)}+\Eb-\Mb+\Thetab^{(k)} \|_F^2\label{eq:proc1}\\
\Xb^{(k+1)} &=& \arg\min_\Xb \frac{\beta}{2} \| \Xb+\Eb^{(k+1)}-\Mb +\Thetab^{(k)} \|_F^2 +  \frac{\mu}{2}  \|\hank\{\Xb\}- \Ub^{(k)}\Vb^{(k)H}+\Lambdab^{(k)}\|^2_F \label{eq:proc11} \\
 \Ub^{(k+1)} &=& \arg\min_\Ub  \frac{1}{2} \|\Ub\|_F^2 + \frac{\mu}{2}  \|\hank\{\Xb^{(k+1)}\}- \Ub\Vb^{(k)H}+\Lambdab^{(k)}\|^2_F \\
  \Vb^{(k+1)} &=&\arg\min_\Vb  \frac{1}{2} \|\Vb\|_F^2 + \frac{\mu}{2}  \|\hank\{\Xb^{(k+1)}\}- \Ub^{(k+1)}\Vb^{H}+\Lambdab^{(k)}\|^2_F \\
   \Thetab^{(k+1)}&=&\Xb^{(k+1)} +\Eb^{(k+1)}-\Mb +\Thetab^{(k)}\\
  \Lambdab^{(k+1)} &=& \hank\{\Xb^{(k+1)}\}- \Ub^{(k+1)}\Vb^{(k+1)H}+\Lambdab^{(k)}
 \end{eqnarray}
 It is easy to show that the first step can be simply reduced to a single soft-thresholding in image patch domain rather than in a lifted Hankel matrix space:
 \begin{gather}\label{eq:proc_sth}
 \Eb^{(k+1)}=\mathcal{S}_{\tau/\beta}\left( \Mb-\Xb^{(k)}-\Thetab^{(k)} \right)
 \end{gather}
 where ${\mathcal S}_\tau$ denotes the pixel by pixel soft-thresholding with the threshold value $\tau$.
 The simple thresholding step  in \eqref{eq:proc_sth} is the main motivation %performed  image patch rather than lifted Hankle matrix,
 which is why we prefer $(P')$ over $(P)$.
 Now, the second step becomes 
 \begin{gather}
 \Xb^{(k+1)}  = 
\frac{1}{\mu+\beta} \left( \mu\hank^{\dag}\left\{ \Ub^{(k)}\Vb^{(k)H} -\Lambdab^{(k)} \right\} -\beta \left( \Eb^{(k+1)}-\Mb +\Thetab^{(k)} \right) \right) ,\label{eq:proc2}
 \end{gather}
 where % $\mathcal{S}_{a}$ is a soft-thresholding operation defined by $\mathcal{S}_{\tau}\left( \cdot \right)=sign\left(\cdot\right)max\{ abs(\cdot)-\tau,0 \}$ and
  $\hank^{\dag}$ corresponds to the Penrose-Moore pseduo-inverse mapping from our block Hankel structure to a patch, which is calculated as
 \begin{equation}
 \hank^{\dag} = \left(\hank^*\hank\right)^{-1}\hank^* \ .
 \end{equation}
Note that the adjoint operator $\hank^*\{\Ab\}$  adds multiple elements of $\Ab$ and put it back to the patch coordinate,
and  $\left(\hank^*\hank\right)^{-1}$ denotes the division by the number of multiple correspondence; hence, the role of the pseudo-inverse is taking the average value and put it back
to the patch coordinate.
Next, the subproblems for $\Ub$ and $\Vb$ can be easily calculated by taking the derivative with respect to each matrix. For example, the derivative of cost function for $\Ub$ is given by
  \begin{eqnarray}
  \frac{\partial L}{\partial \Ub}&=&\frac{\partial }{\partial \Ub}\left(  \frac{1}{2} \|\Ub\|_F^2 +  \frac{\mu}{2}  \|\hank\{\Xb\}- \Ub\Vb^H+\Lambdab\|^2_F \right)\nonumber\\
	&=&\Ub-\mu \left( \hank\{\Xb\}-\Ub\Vb^H+\Lambdab \right)\Vb\nonumber\\
	&=&\Ub\left( I+\mu\Vb^{H}\Vb \right)-\mu \left( \hank\{\Xb\}+\Lambdab \right)\Vb,\nonumber
\end{eqnarray}
and the closed-form solution of the subproblem for $\Ub$ is obtained by setting $\partial L/\partial \Ub=0$.
In the similar way, the derivative with respect to $\Vb$ can be obtained. Accordingly, the closed-form update equations for $\Ub$ and $\Vb$ are given by
\begin{eqnarray}
 \Ub^{(k+1)}  &=& \mu \left(\hank\{\Xb^{(k+1)}\}+\Lambdab^{(k)}\right)\Vb^{(k)} \left(I+\mu \Vb^{(k)H}\Vb^{(k)}\right)^{-1} \label{eq:U}  \\
  \Vb^{(k+1)} &=& \mu\left(\hank\{\Xb^{(k+1)}\}+\Lambdab^{(k)}\right)^H\Ub^{(k+1)} \left(I+\mu \Ub^{(k+1)H}\Ub^{(k+1)}\right)^{-1}   \label{eq:V}.
 \end{eqnarray}
Even though the original Hankel matrix  $\hank\{\Xb\}$ has large dimension,  it is important to note that our algorithm using \eqref{eq:U} and \eqref{eq:V} 
only require the matrix inversion of $k\times k$ matrix, where $k$ denotes the estimated rank of the Hankel matrix. This significantly reduces the overall computational complexity.

Before we apply ADMM, the initial estimatel $\Ub$ and $\Vb$ have to be determined with an estimated rank. For this, we employed an SVD-free algorithm called low-rank factorization model (LMaFit) \cite{wen2012solving}.
 More specifically, for a low-rank matrix $\Zb$, LMaFit solves the following optimization problem:
\begin{equation}\label{eq:lmafit}
\min_{\Ub,\Vb,\Zb} ~\frac{1}{2} \|\Ub\Vb^H-\Zb\|^2_F~% \mbox{subject to}  \quad \Zb_{i,j} = \Mb_{i,j},
\end{equation}
and $\Zb$ is initialized with $\hank\{\Mb\}$.
LMaFit solves a linear equation with respect to  $\Ub$ and $\Vb$ to find their updates and relaxes the updates by taking the average between the previous iterations.
Moreover, the rank update can be done automatically by detecting the abrupt changes of diagonal elements of QR factorization \cite{wen2012solving}. Even though the problem \eqref{eq:lmafit} is non-convex due to the multiplication of $\Ub$ and $\Vb$,
the convergence of LMaFit to a stationary point was analyzed in detail \cite{wen2012solving}. However, the LMaFit alone cannot recover the block Hankel structure, which is the reason we use ADMM later to impose the structure. 

%In our application,  the image patch is corrupted with impulse noise, so the tolerance for rank estimation should be reduced by considering the noise.

%%\subsection{Image Partitioning}
%
%\begin{figure}[!ht]
%\centering
%\includegraphics[width=\imsize in]{./figures/Drawing1_sparse&lowrank.png}\vspace{-0.3cm}
%\caption{The reconstruction flowchart of the proposed sparse and low-rank using patch-based Hankel approach for removal of impulse noise.}
%\label{fig:flowchart}
%\end{figure}

\section{Extension to Multi-channel Impulse Noise Removal}\label{sec:mul}
\label{sec:color}

In many applications, images are obtained through multiple measurement channels. For example, in a colour image,  
multiple images are measured throughout R (red), G (green) and B (blue) detectors. In multispectral imaging for remote sensing applications, a scene
is measured through many  spectral bands.
In these applications,  % even though each measurement step through respective channel is independent to each other, 
the underlying structure is identical so that there exists strong
correlation between different channel measurements.
Regarding the random impulse noise contamination, we may encounter two different scenarios: 1) noisy pixel locations are independent between the channels,
and 2) noisy pixel locations are same across the channels.  The first scenario is commonly observed when independent detectors are used for each channel. %, where  the
%faulty detector locations are independent at each channel.  
On the other hand, when a spectrometer is used to split an input into multiple channels, then the noisy pixel locations should be common across the channel. % each other.
Therefore, in this section, we are interested in  extending  single-channel robust ALOHA to address these two cases.

\subsection{Multi-Channel Image Modeling}

\begin{figure}[!bt]
\centering
\includegraphics[width=15cm]{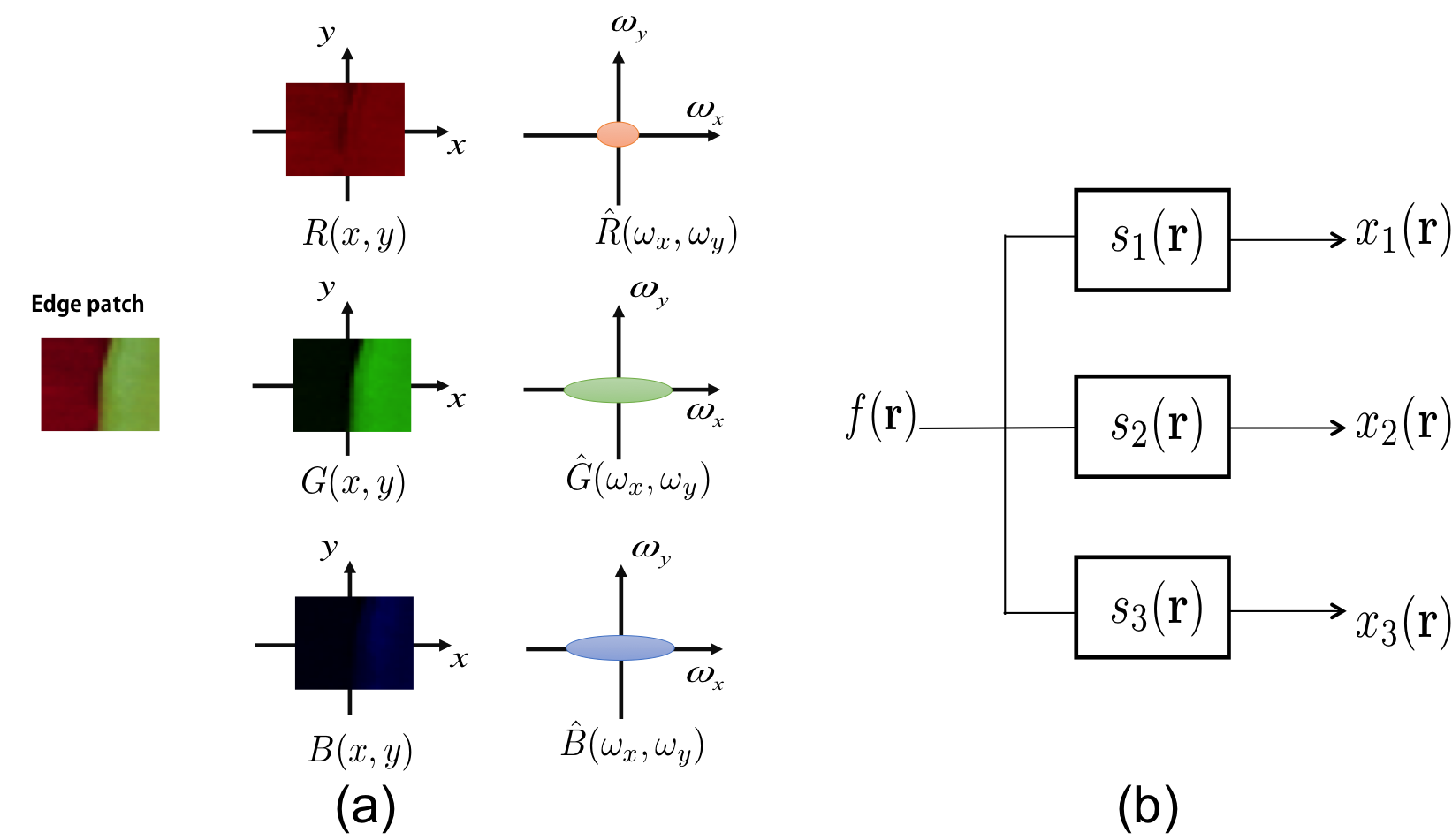}
\caption{(a) Spectral distribution across channels.  (b) Multi-channel measurement model.}
\label{fig:spectral_patch}
\end{figure} 

Let $f(\rb)$ denote an underlying image patch that are common for all channel measurements,
 and $\hat f(\omegab)$ be its spectrum. 
Then,  as shown in Fig.~\ref{fig:spectral_patch}(a),  the spectrum of the $i$-th channel measurement can be modeled as: 
\begin{equation}\label{eq:model}
\hat x_i(\omegab) = \hat s_i(\omegab) \hat f(\omegab),\quad i=1,\cdots, C, 
\end{equation}
where $\hat s_i(\omegab)$ denotes a spectral modulation function of the $i$-th channel.
The model \eqref{eq:model} assumes that each channel measurements still retains the textures of the
underlying images with a channel specific modulation, which property was previously extensively used in multichannel deconvolution problems  \cite{harikumar1999exact,harikumar1999perfect,harikumar1998fir} illustrated in
Fig.~\ref{fig:spectral_patch}(b).
As a result,  it is easy to derive the following 
%inter-coil relation in image domain:
%\begin{eqnarray*}
%s_j(\rb)  \ast  x_i(\rb) = s_i(\rb) \ast x_j (\rb), \quad i,j=1,\cdots, C, \mbox{ and } i\neq j,
%\end{eqnarray*}
%which provides the following
 {\em inter-channel annihilating filter relation}:
\begin{eqnarray}\label{eq:an_mat2}
s_j(\rb)  \ast  x_i(\rb) - s_i(\rb) \ast x_j (\rb)= 0 , \quad \forall \rb , ~~ i\neq j,
\end{eqnarray}
which  was also a key property in these multichannel devolution algorithms \cite{harikumar1999exact,harikumar1999perfect,harikumar1998fir}.

To exploit the inter-channel annihilating property in our robust ALOHA, we construct the following matrix:
\begin{equation}\label{eq:hank_mul}
	\Ybc=\left[ \hank\{ \Xb_{1} \}~ \hank\{ \Xb_{2} \} ~\cdots~ \hank\{ \Xb_{C} \} \right]  ~\in \Rd^{(N-q+1)(M-p+1) \times Cpq}
\end{equation}
where $\hank\{\Xb_i\}$ denotes the Hankel structured matrix constructed from the $i$-th channel measurement $x_i(\rb)$.
Then, it is easy to show that
$$\Ybc \Sc_1 =\zerob,$$ 
where $\Sc_1$ is defined recursively as follows:
\begin{eqnarray}
  \Sc_{C-1} &\triangleq& \left[
                  \begin{array}{cc}
                    \bar\sb_C \\    - \bar\sb_{C-1}  \\
                  \end{array}
                \right]
   \\
  \Sc_t &\triangleq&\left[
                      \begin{array}{cccc|c}
                       \bar\sb_{t+1} & \bar\sb_{t+2} & \cdots & \bar\sb_{C}  & \zerob \\ \hline
                       -\bar\sb_t &  & && \\
                         & -\bar\sb_t &  &  & \Sc_{t+1}  \\
                          &  & \ddots &  \\
                         &  &  &   -\bar\sb_t  & 
                      \end{array}
                    \right]  \ ,
 \end{eqnarray}
and
$\bar\sb_i := \overline{\textsc{vec}(\Sb_i)}$
denotes the reversed ordered, vectored spectral modulation filter for the $i$-th channel.
Because
$ \dim \Null(\Ybc)= \mathrm{rank}(\Sc_1) = \binom{C}{2} = {C(C-1)}/{2}, $
we have
\begin{eqnarray}
\mathrm{rank}~ \Ybc  &\leq&  k C-  \frac{C(C-1)}{2} 
=
\frac{C(2k-C+1)}{2}  \ ,
\end{eqnarray}
when the underlying spectrum $\hat f(\omegab)$ is $k$-sparse.
%This concludes the proof.
Hence, by choosing the annihilating filter size sufficiently large, we can make $\Ybc$  low-ranked
and  the aforementioned  sparse+ low rank model can be used for impulse noise removal.

\subsection{Optimization Methods}

\subsubsection{Channel independent impulse noises}

When the locations of the impulse noise are  independent between  channels,  then the associate optimization problem is very similar to that of the single channel problem.
More specifically, if we define 
\begin{equation}
 \Mb= \begin{bmatrix} \Mb_1 | \cdots | \Mb_C \end{bmatrix},\quad  \Xb= \begin{bmatrix} \Xb_1 | \cdots | \Xb_C \end{bmatrix}, \quad  \Eb= \begin{bmatrix} \Eb_1 | \cdots | \Eb_C \end{bmatrix}
\end{equation}
such that the each channel measurement is give by $$\Mb_i=\Xb_i+\Eb_i,$$
then  the optimization problem  becomes
\begin{eqnarray}
\min_{\Eb,\{(\Ub,\Vb)| \Ub\Vb^H =\Ybc\}}  && \|\Ub\|_F^2+ \|\Vb\|_F^2 +\tau\|\Eb\|_1 \label{eq:fac2} \\
\mbox{subject to} %&&\hank\{\Xb\} = \Ub\Vb^H \label{eq:fac}\\
&&  \Ybc = \left[ \hank\{ \Xb_{1} \}~ \hank\{ \Xb_{2} \} ~\cdots~ \hank\{ \Xb_{C} \} \right] \notag\\
&& \Xb+\Eb=\Mb \ . \label{eq:data2}
\end{eqnarray}
In this case,  the Lagrangian cost function  and the associated 
 subproblems   are same as  Eq. \eqref{eq:ADMM}  and   \eqref{eq:proc11}-\eqref{eq:proc2}, respectively. % in the single-channel case.

\subsubsection{Common impulse noise locations}

When the noisy pixel locations are common across the channels, we need a major algorithmic change that  comes  from a common
sparsity inducing matrix norm penalty.  More specifically, a common support condition of sparse components $\Eb$ should be imposed across channels.
In this paper, this constraint is  formulated using the group-wise mixed $l_{1,2}$ norm:
\begin{equation}\label{eq:yc}
 \Eb= \begin{bmatrix} \Eb_1 | \cdots | \Eb_C \end{bmatrix}  \quad    \Longrightarrow \quad \|\Eb\|_{1,2}:=\sum_{i,j}\sqrt{\sum_{k=1}^C \Eb_k(i,j)^2} 
\end{equation} 
Then,  Eq. \eqref{eq:ADMM} can be converted to
\begin{eqnarray}\label{eq:ADMM2}
L(\Ub,\Vb,\Eb,\Xb,\Thetab,\Lambdab) :=&& \frac{1}{2} \left( \|\Ub\|_F^2+\|\Vb\|_F^2\right) + \tau \|\Eb\|_{1,2}+\frac{\beta}{2} \| \Xb+\Eb-\Mb+\Thetab \|_F^2  + \nonumber\\
 && \frac{\mu}{2}  \|\hank\{\Xb\}- \Ub\Vb^H+\Lambdab\|^2_F.
\end{eqnarray}
Then, instead of  using \eqref{eq:proc1},
the corresponding subproblem for $\Eb$ has the following closed form  solution \cite{ramani2011parallel}:
 \begin{eqnarray}
 \Eb^{(k+1)} &=& \mathcal{S}^{{ch}}_{\tau/\beta}\left( \Mb-\Xb^{(k)}-\Thetab^{(k)} \right) 
 \end{eqnarray}
where, for $\Eb$ in \eqref{eq:yc},  $ \mathcal{S}^{{ch}}_\lambda(\Eb)$ is defined as
$$
\mathcal{S}^{{ch}}_{\lambda}\left(\Eb\right) : =  \begin{bmatrix}  \mathcal{S}^{{vec}}_{\lambda}(\Eb_1) | \cdots |  \mathcal{S}^{{vec}}_{\lambda}(\Eb_C) \end{bmatrix}
$$
and
\begin{equation}
  \left[\mathcal{S}^{{vec}}_{\lambda}(\Eb_k) \right]_{ij} =    \frac{\Eb_k(i,j)}{\sqrt{\sum_{k=1}^C|\Eb_k(i,j)|^2}} \max\left\{ \sqrt{\sum_{k=1}^C|\Eb_k(i,j)|^2}-\lambda,0\right\} \ .
\end{equation}
 Other remaining subproblems   are same with   \eqref{eq:proc11}-\eqref{eq:proc2}.

\section{Experimental Results}\label{sec:result}
\subsection{Removal of Random Valued Impulse Noise (RVIN)}\label{sec:texture}

We first performed denoising experiments using randomly distributed random valued impulse noise (RVIN)  that corrupts  25\% and 40\% of the whole  image pixels.
The RVIN is given as follows. Let $x_{ij}:=\Xb(i,j)$ and $N(x_{ij})$ be the original pixel value at location $(i,j)$ and the contaminated pixel with impulse noise at location $(i,j)$, respectively. When the dynamic range of pixel value is given as $\left[ d_{{min}}~ d_{{max}}\right]$, RVIN is described as
\begin{equation}\label{eq:rvin}
N(x_{ij})=
\begin{cases}
d_{ij}       & \quad \text{ with probability $p$}\\
x_{ij}  & \quad \text{ with probability $1-p$}\end{cases}
\end{equation}
where $d_{ij}$ is a random number between $\left[ d_{{min}}~ d_{{max}}\right]$ chosen by the uniform random probability density function and $p$ is the proportion of noisy pixels with respect to total pixels.

 The test sets were Baboon, Barbara, Boat, Cameraman, House, Lena and Peppers images. All test images were rescaled to have values between 0 and 1. 
For comparison, a median filter method (MATLAB built-in function `medfilt2', indicated as MF in the figures) was used as the simplest reference algorithm,
and the existing algorithms such as ACWMF\cite{chen2001adaptive}, and TVL1\cite{getreuer2012rudin}  were also used.  The original codes from the original authors were used and
the parameters for the comparative algorithms were optimized to have the best performances. 
The parameters for the proposed method are given  in Table. \ref{tab:param}. The maximum iteration number of ADMM in Eq. \eqref{eq:ADMM} was set to $500$,
 and the stopping criteria was defined as in \cite{wen2012solving}
with the tolerance  set to $10^{-4}$.
For quantitative evaluation, we used the PSNR (peak signal-to-noise ratio). Specifically, when the reference signal ($\yb$) is given, the PSNR of the reconstructed image ($\xb$) is calculated as
\begin{equation*}
\mbox{PSNR}(\xb)=20\log_{10} \left( \frac{\|\yb\|_\infty}{1/\sqrt{N}\times \|\yb-\xb\|_2} \right). 
\end{equation*}
%PSNR becomes higher when the quality of image is enhanced. 

\begin{table}[!h]\scriptsize
\centering\label{tab:param}
\caption{ Hyper-parameters used in the proposed algorithm. $\mu$ and $\beta$ of ADMM are $1$. For LMaFit, initial rank with increasing strategy is one and tolerances for $25\% (40\%)$ impulse noise is $0.2 (0.3)$.} \label{tab:parameter}
\renewcommand{\arraystretch}{1.2}
\begin{tabular}{l||c|c|c|c|c|c|c}
  \hline\hline
  		& Baboon  	& Barbara  	& Boat  	& Cameraman & House & Lena & Peppers\\
	& ($512\times512$) & ($512\times512$) & ($512\times512$) &  ($256\times256$) &  ($256\times256$) &($512\times512$)& ($255\times255$) \\
\hline
\hline
Size of $\Xb$	& $45\times45$ &$25\times25$&$25\times 25$		& $31\times 31$&$25\times 25$			& $25\times25$ 	& 	$25\times25$ ($45\times45$)\\
\cline{1-8}
Size of $\Hb$	& $13\times13$ &$11\times11$&$11\times 11$		& $13\times 13$&$11\times 11$			& $11\times11$ 	& 	$9\times9$ ($13\times13$) \\
\cline{1-8}
Size of $\hank\{\Xb\}$ & $1089\times169$&$225\times 121$&$225\times 121$	& $361\times169$&$225\times121$			& $225\times 121$ 	& 	$289\times81 (1089\times169)$ \\
\cline{1-8}
$\tau (\times 10^{-2})$ 				& $10~(7.5)$    	& $10$&$10$		& $10~(7.5)$	& $10$	& $10$ 	& 	$10~(7.5)$ \\
%\cline{1-8}
%$Tol for LMaFit (\times 10^{-1})$ 	& $2,~(3)$     	& $2,~(3)$&$2,~(3)$		& 1	& 10			& 10 	& 	1 \\
\hline\hline
\end{tabular}
\end{table}
\normalsize

\begin{table}[!htb]
\caption{Reconstruction image PSNR by various denoising algorithms from 25\% and 40\% random valued impulse noise (RVIN). The highest PSNR in each image are highlighted with boldface.}\label{tab:psnr}% The averaged gains of inpainting algorithms are calculated relatively with Mesh interpolation.
\centering
\begin{tabular}{c|c||ccccc}
  \hline\hline
\multicolumn{2}{c||}{\diagbox{Test\\Images}{Algorithms}}	& Noisy image  &MF  	& ACWMF  	& TVL1  & {Proposed}\\
\hline\hline
  \multirow{2}{*}{Baboon}	&25\% 	&  15.53	& 22.01	 &  23.14	&  23.39   &    \textbf{24.82} 	\\
  				    	&40\% 	& 13.52     &  20.52	&   21   &  21.68 	&    \textbf{22.46} 	\\
\hline
\multirow{2}{*}{Barbara}	&25\% 	& 14.84  	& 23.62	 & 24.59 	& 24.99    & \textbf{33.13} 	\\
 					&40\% 	& 12.82  	& 21.14     & 22.86  	& 23.43 	&\textbf{28.51} 	\\
\hline
 \multirow{2}{*}{Boat} 	&25\% 	& 15.37 	& 27.27 	& 28.09 	& 28.64 	& \textbf{30.57}\\
       					&40\% 	& 13.31 	& 23.52 	& 25.73 	& 26.19	&  \textbf{26.89}\\
\hline
  \multirow{2}{*}{Cameraman}&25\% 	& 14.44 	& 24.28	& 24.69 	& 25.58 	 &\textbf{26.74} 	\\
       					&40\%  	& 12.37 	& 20.61	& 22.13 	& 23 	 	& \textbf{23.44}\\
\hline
  \multirow{2}{*}{House}   	&25\% 	& 15.31 	& 29.08 	& 29.74 	& 31.16 	&\textbf{34.29} 	\\
       					&40\% 	& 13.27 	& 24.36 	& 27.44 	& 28.07	& \textbf{28.99}\\
\hline
  \multirow{2}{*}{Lena} 	&25\% 	& 15.29 	& 30.07 	& 30.75 	& 31.9 	& \textbf{34.31} 	\\
       					&40\%  	& 13.23 	& 24.6 	& 28.78 	& 29.14	& \textbf{30.48} 	\\
\hline
  \multirow{2}{*}{Peppers}  &25\% 	& 14.99 	&27.47 	& 27.91 	& 29.09 	&\textbf{29.47} 	\\
       					&40\% 	& 12.96 	& 23.15 	& 25.66 	& 26.07 	& \textbf{26.26}\\
\hline
\hline
\end{tabular}
\end{table}

% x5 down sampled Barbara
\begin{figure}[!hbt]
\centering
\includegraphics[trim = 0mm 18mm 0mm 9mm,clip=true,width=\imsizes in]{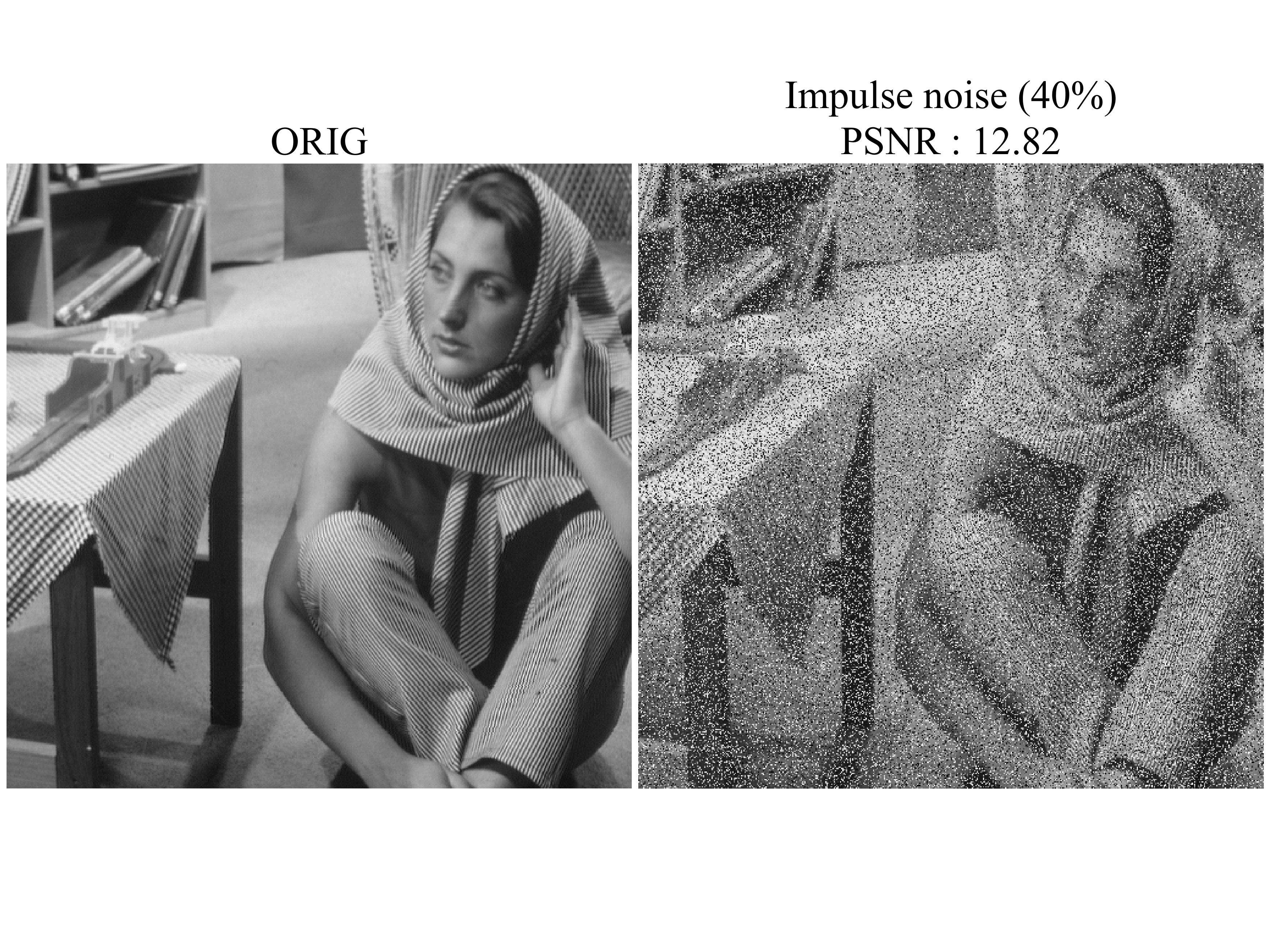}
\includegraphics[trim = 0mm 18mm 0mm 9mm,clip=true,width=\imsizes in]{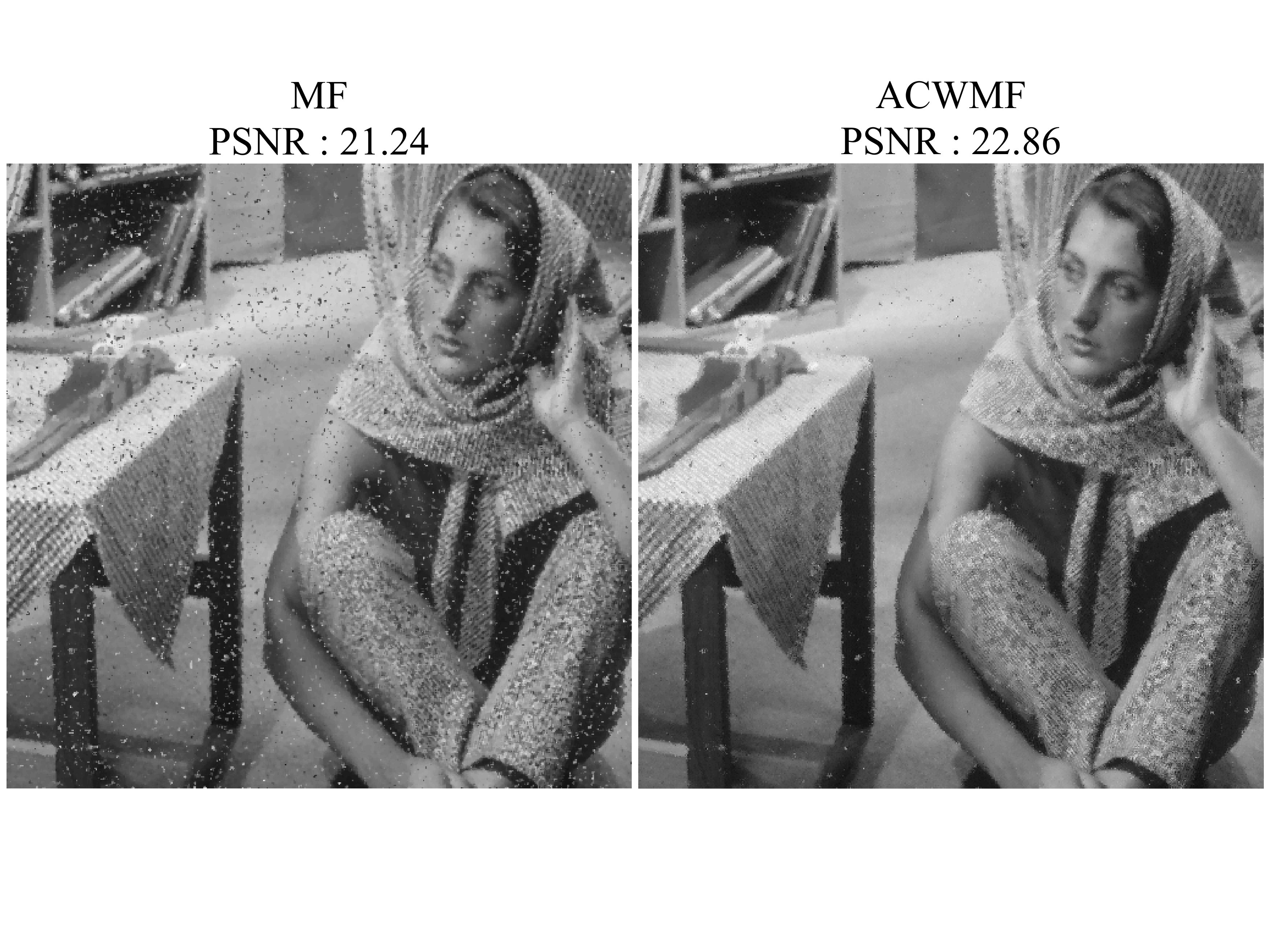}
\includegraphics[trim = 0mm 18mm 0mm 9mm,clip=true,width=\imsizes in]{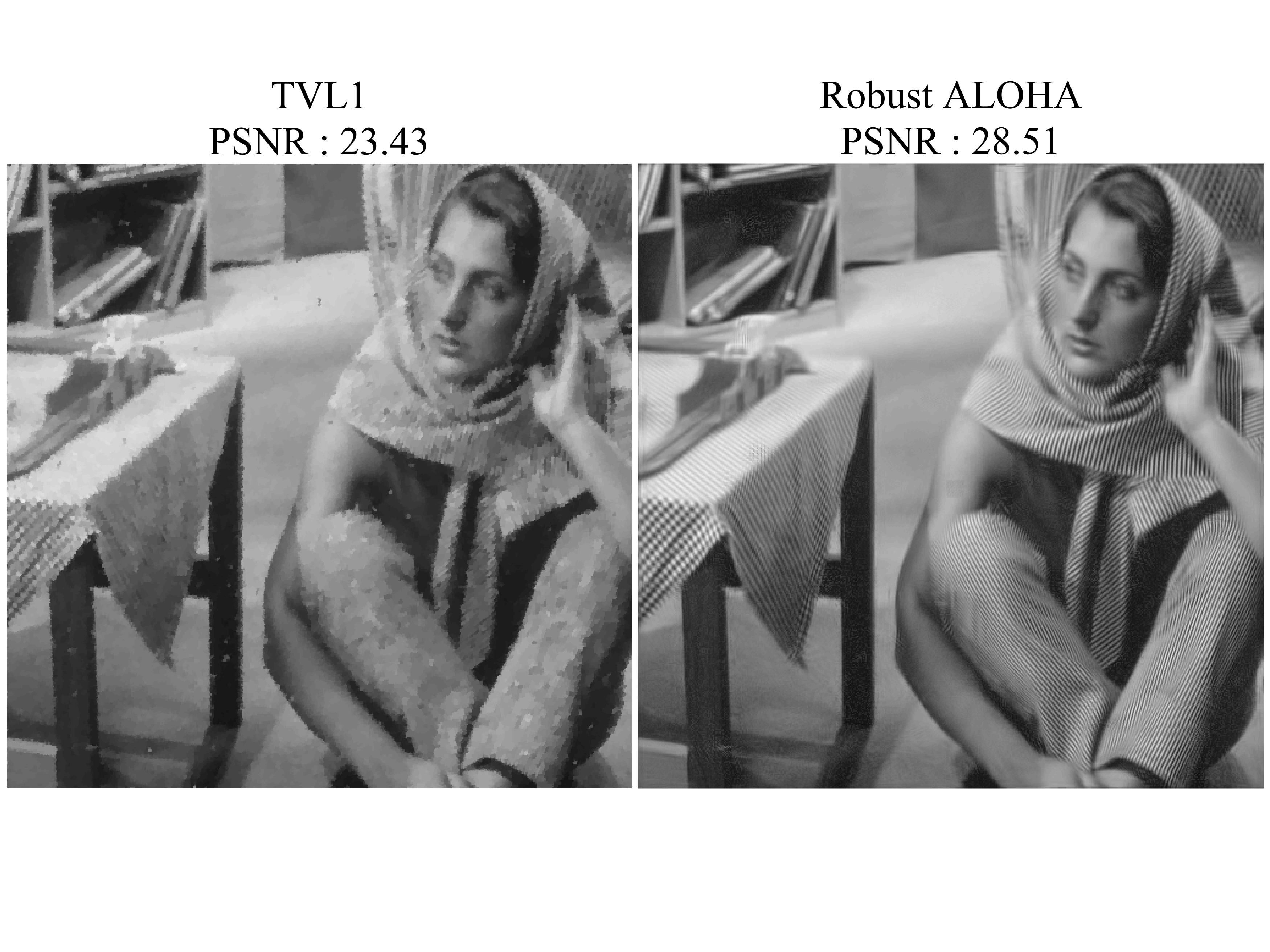}
\caption{Reconstructed Barbara images by various methods from 40\% random valued impulse noise.}
\label{fig:result_barbara_x5}
\end{figure} 

\begin{figure}[!hbt]
\centering
\includegraphics[trim = 0mm 18mm 0mm 12mm,clip=true,width=\imsizes in]{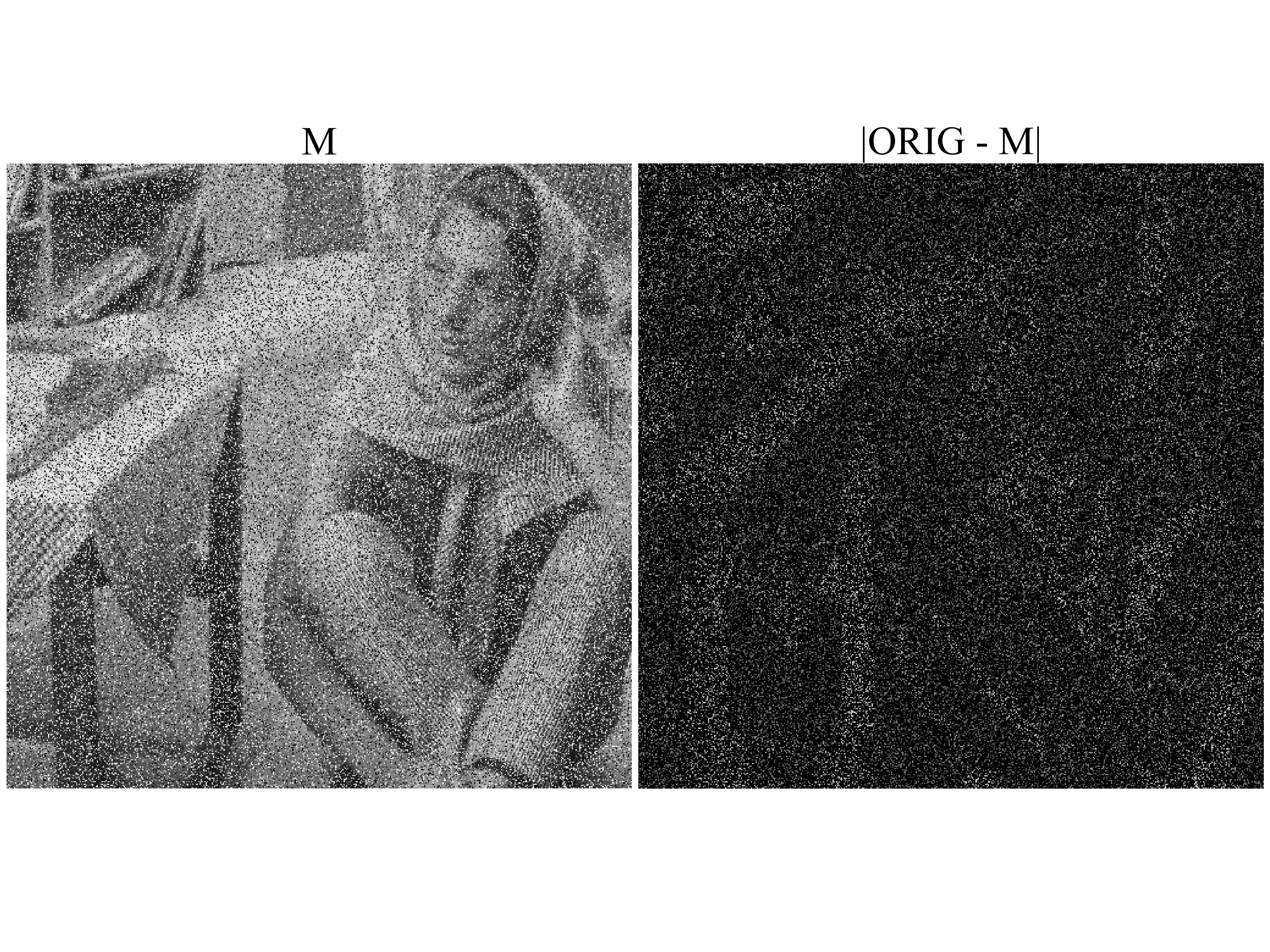}
\includegraphics[trim = 0mm 18mm 0mm 12mm,clip=true,width=\imsizes in]{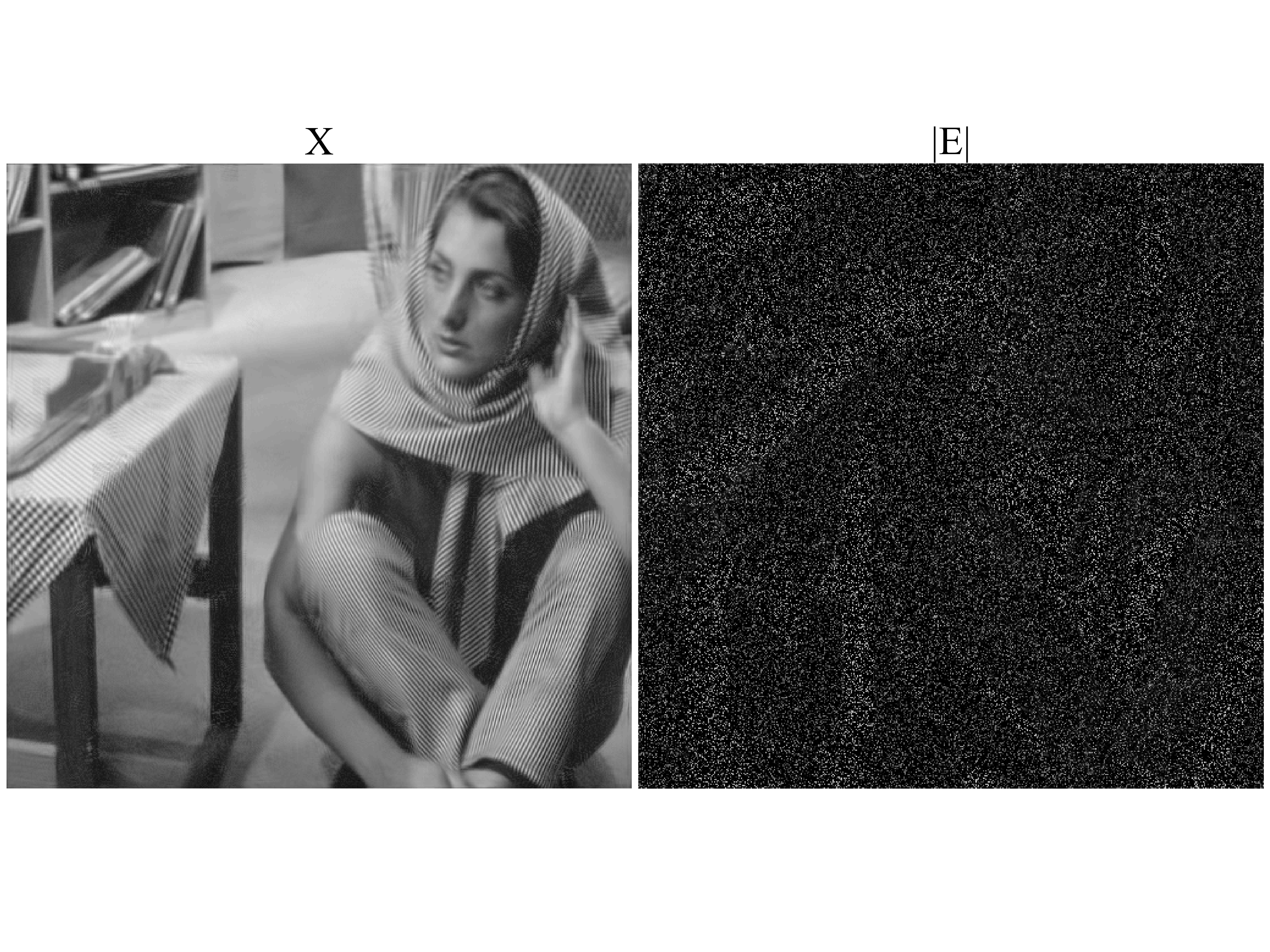}
\caption{$\Mb$: the measured noisy image. $|\mbox{ORIG}-\Mb|$: the residual image between noisy image and original image. $\Xb$: the decomposed low-rank part, and $\Eb$: the decomposed
 sparse component.}
\label{fig:res_barbara_mxe}
\end{figure} 

We summarized the PSNR results  in Table \ref{tab:psnr} for all reconstructed images. %are shown in Figs.~\ref{fig:result_barbara_x5}. 
The proposed method outperformed all other  algorithms in both PSNR and visual quality, and the PSNR improvement was up to 8dB.  %Barbara 
Fig.~\ref{fig:result_barbara_x5}  is the typical reconstruction result  that shows noticeable enhancement  in both visual quality and quantitative measures. 
In order to show that the proposed sparse+ low rank decomposition properly decomposes the impulse noises from the images,
the decomposed sparse and low-rank components are illustrated   in Fig. \ref{fig:res_barbara_mxe}. We can see that sparse component ($|\Eb|$) looks similar to the additive impulse noises as indicated by $|\mbox{ORIG}-\Mb|$.

\subsection{Salt and Pepper Noise}

The salt and pepper noise is a special case of RVIN, because it has impulse noises with the intensity of the minimum and maximum values of pixel
dynamic range. Specifically, the salt and pepper noises are given by
\begin{equation}
N(x_{ij})=
\begin{cases}
d_{min}       & \quad \text{ with probability $p/2$}\\
d_{max}       & \quad \text{ with probability $p/2$}\\
x_{ij}  & \quad \text{ with probability $1-p$}\\
\end{cases}
\end{equation}
where variables  $p,~d_{max},~d_{min}$ were defined in Eq. \eqref{eq:rvin}.
\begin{figure}[!h]
\centering
\includegraphics[trim = 0mm 0mm 0mm 0mm,clip=true,width=\imsizes in]{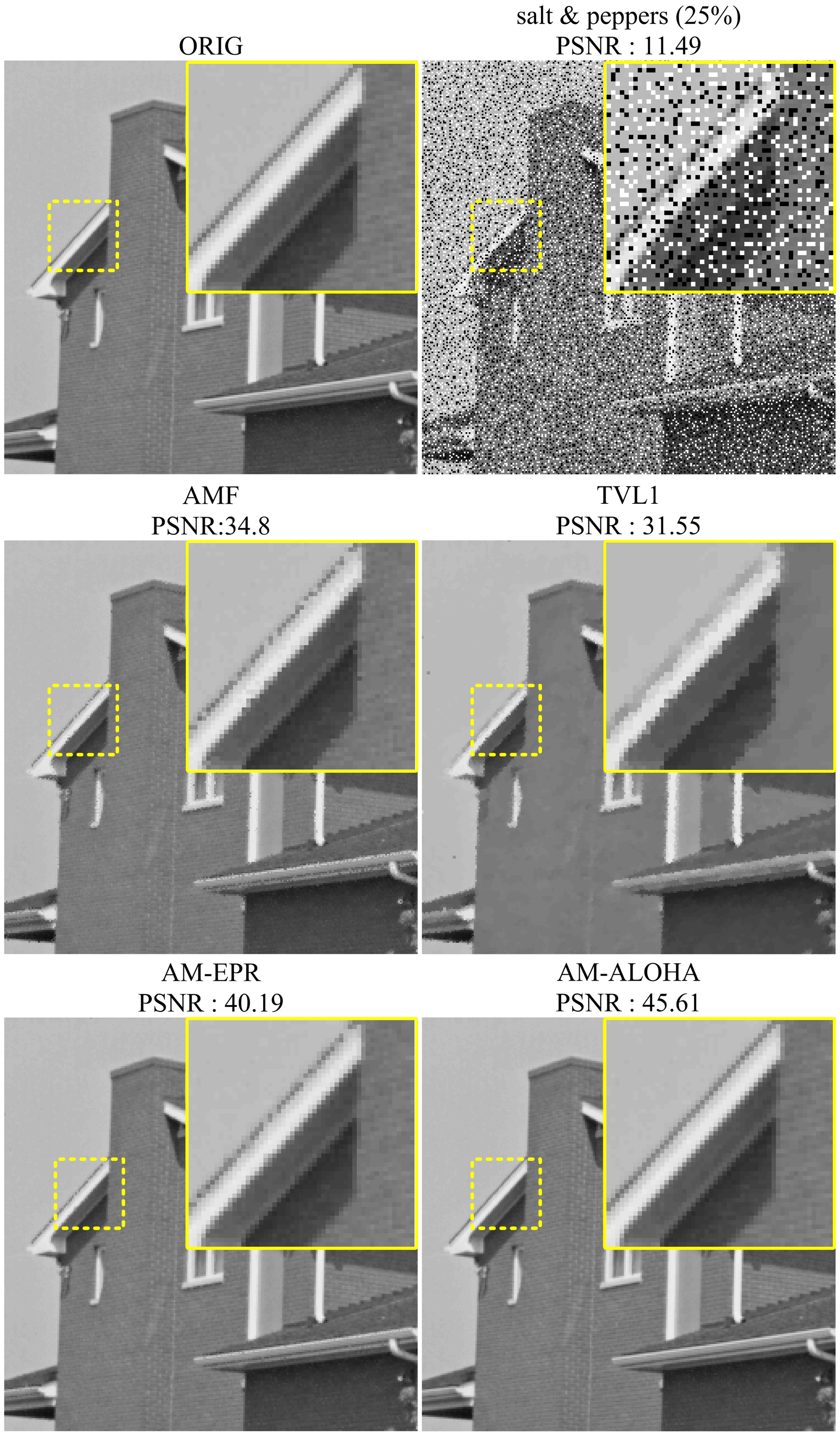}
\caption{Salt and pepper noise removal results using various methods.}
\label{fig:res_house}
\end{figure}
Due to the extremal pixel values,  the salt and pepper noise can be exceptionally well detected by adaptive median filter (AMF), so the AMF-based denoising algorithms with edge preserving prior (AM-EPR) have been proposed \cite{chan2005salt}. 
If the position of  the salt and noise locations are well detected by AMF,  our robust ALOHA can be modified accordingly.
More specifically,  the estimation of the sparse component $\Eb$ in \eqref{eq:fac} and \eqref{eq:data} is no more necessary,
and the resulting algorithm becomes identical to  the ALOHA image inpainting algorithm \cite{7127011}. We denote this modification using AMF as AM-ALOHA.
To demonstrate that our algorithm still outperforms the existing algorithms,  we compared our method with
median filtering (MF), adaptive median filtering (AMF),  AMF-based denoising algorithms with edge preserving prior (AM-EPR)
using 
 25 \% salt and pepper noise.
 The results in Fig.~\ref{fig:res_house} clearly demonstrates that the proposed AM-ALOHA outperforms all other algorithms.

\subsection{Multi-channel Denoising}
% (RVIN) Noisy | TVL1 | ALOHA (Peppers)

\begin{figure}[!hbt]
\centering
\includegraphics[trim = 0mm 0mm 0mm 0mm,clip=true,width=\imsizes in]{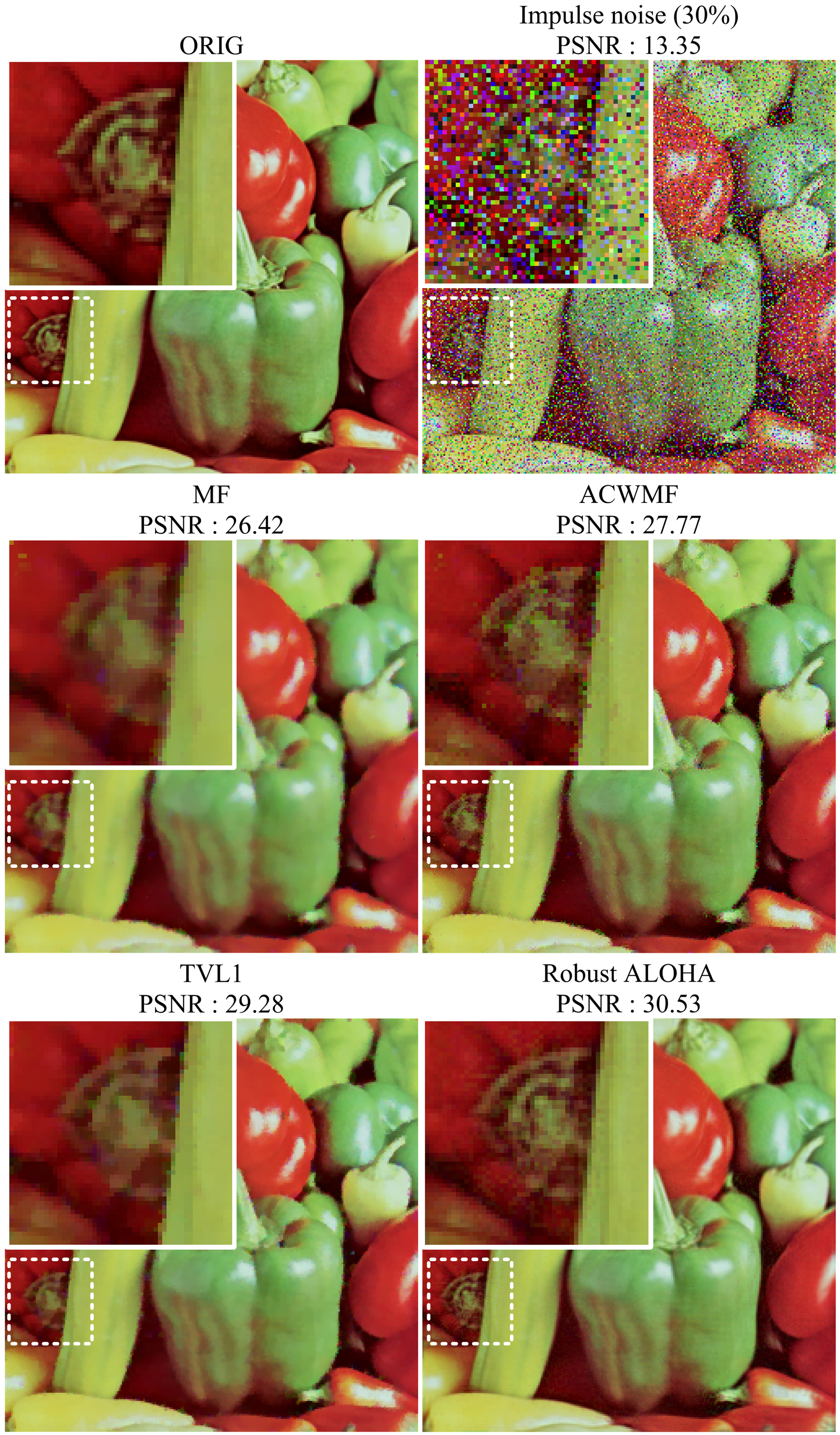}
\caption{Multi-channel denoising results under 30\% random valued impulse noises at the independent pixel locations at each RGB channels.}
\label{fig:rgb_indep}
\end{figure}

\begin{figure}[!hbt]
\centering
\includegraphics[trim = 0mm 0mm 0mm 0mm,clip=true,width=\imsizes in]{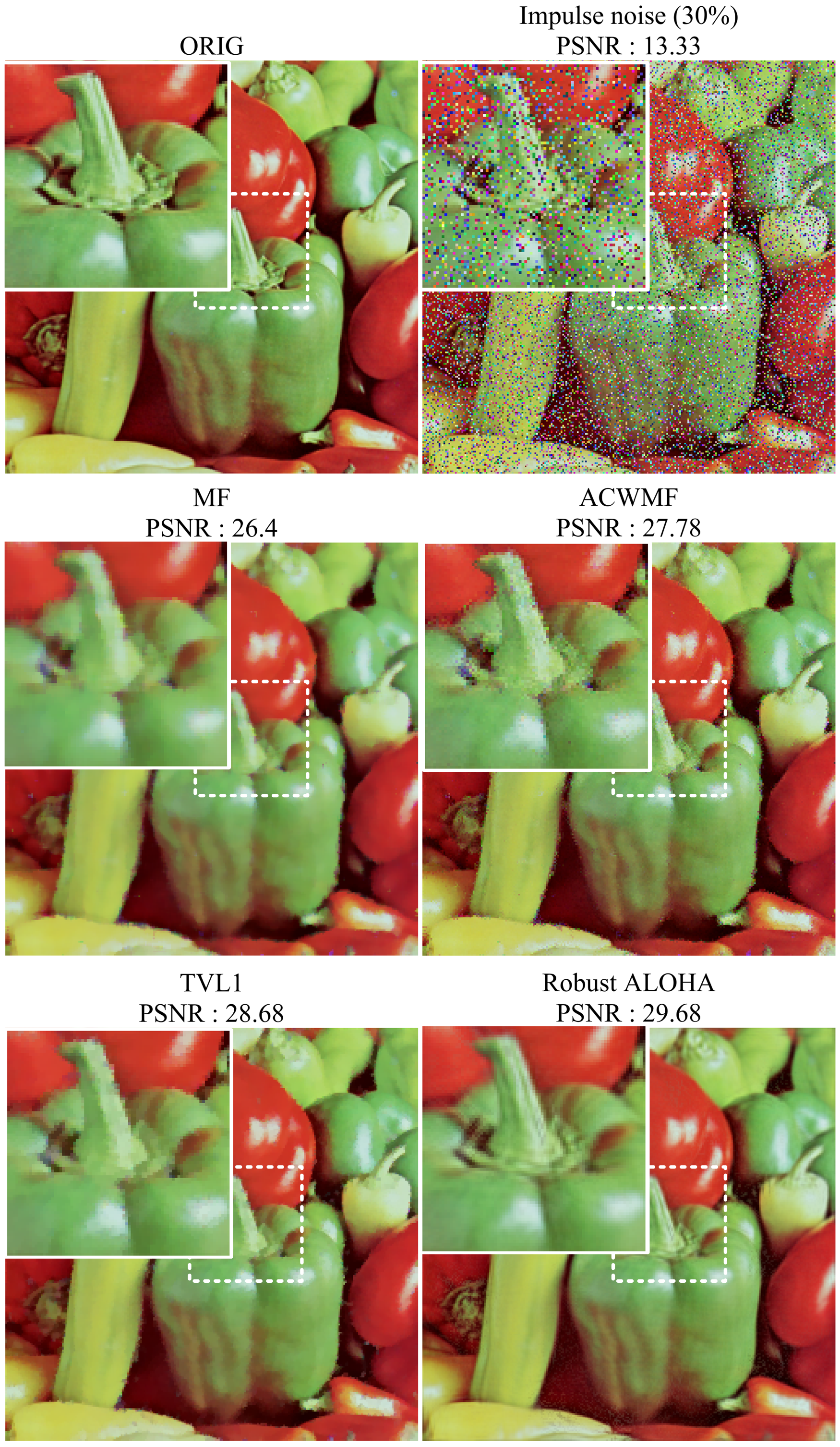}
\caption{Multi-channel denoising results under 30\% random valued impulse noises at the common pixel locations across the RGB channels.}
\label{fig:rgb_co}
\end{figure}

To verify that the proposed method can be easily extended to multichannel images,
we conducted experiments with colour RGB images.
As discussed before, the noisy pixel location can be either same across channels or independent for each channel. So we conducted experiments under the two different scenarios.
 Fig. \ref{fig:rgb_indep} showed the reconstruction result when  $30\%$ channel independent impulse noises were added,
 whereas  
Fig. \ref{fig:rgb_co} corresponds to the scenario when   $30\%$  of impulse noises were added at the same locations across RGB channels.
The proposed method provided better  detailed structures (e.g. bundle of peppers and the edges of peppers) than TVL1 method as shown  in Figs. \ref{fig:rgb_indep}-\ref{fig:rgb_co}. Also, the cartoon-like artifacts were significantly reduced in the proposed method. In the inset images, the detail structures of peppers are magnified to demonstrate  superior performance of the proposed method over TVL1.

One of the interesting observations from these experiments was that the proposed reconstruction provided a better PSNR for the channel independent impulse noises.
This was because the noiseless pixel values from other channels could improve the image inpainting performance  of noisy pixel values by exploiting the correlation between the channels.


\subsection{Comparison with conventional RPCA}
% (RVIN) RPCA | patch-based RPCA | ALOHA 
% (SP) RPCA | patch-based RPCA | ALOHA 
\begin{figure}[!hbt]
\centering
\includegraphics[trim = 0mm 18mm 0mm 8.5mm,clip=true,width=\imsizes in]{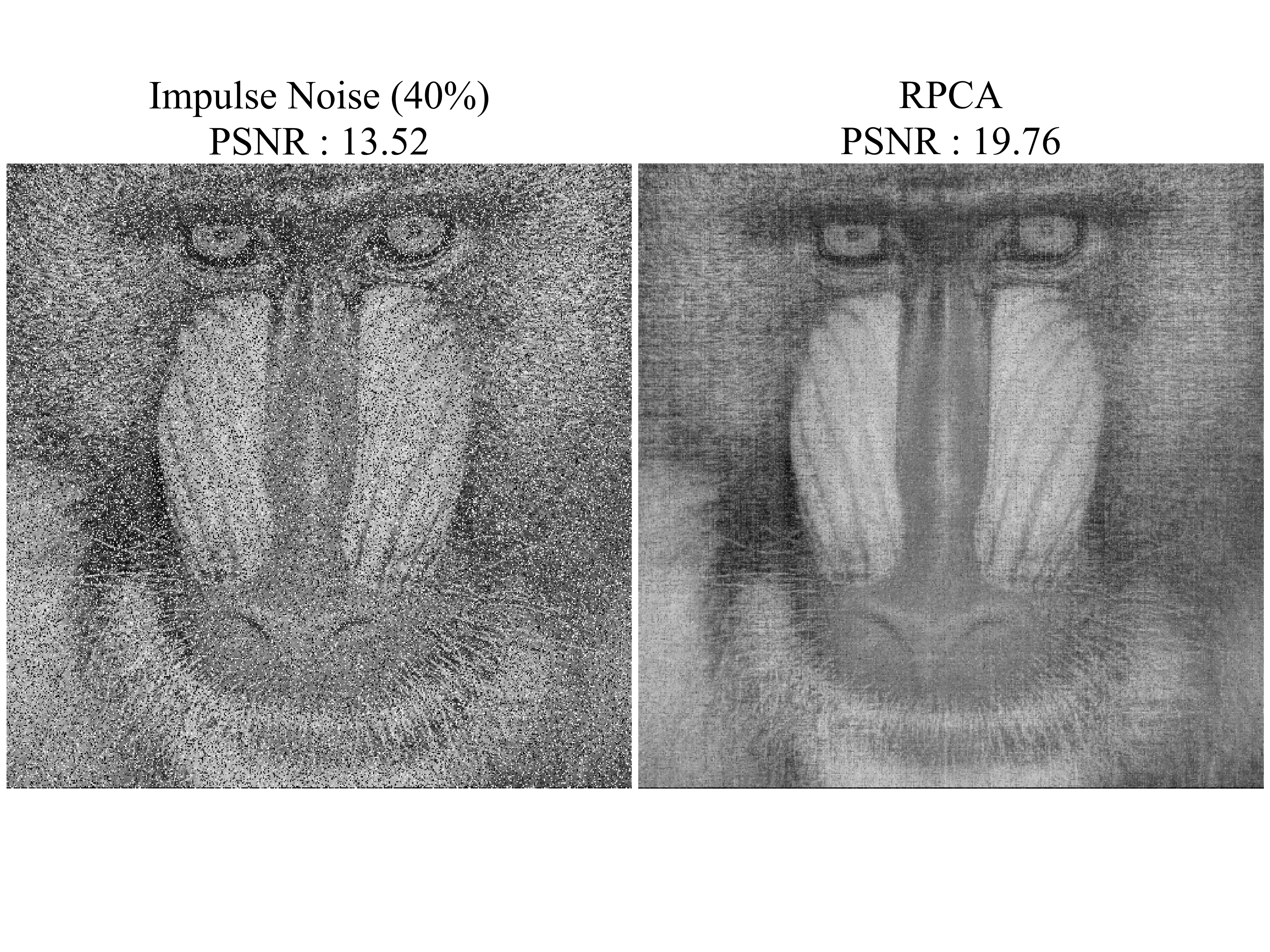}
\includegraphics[trim = 0mm 18mm 0mm 8.5mm,clip=true,width=\imsizes in]{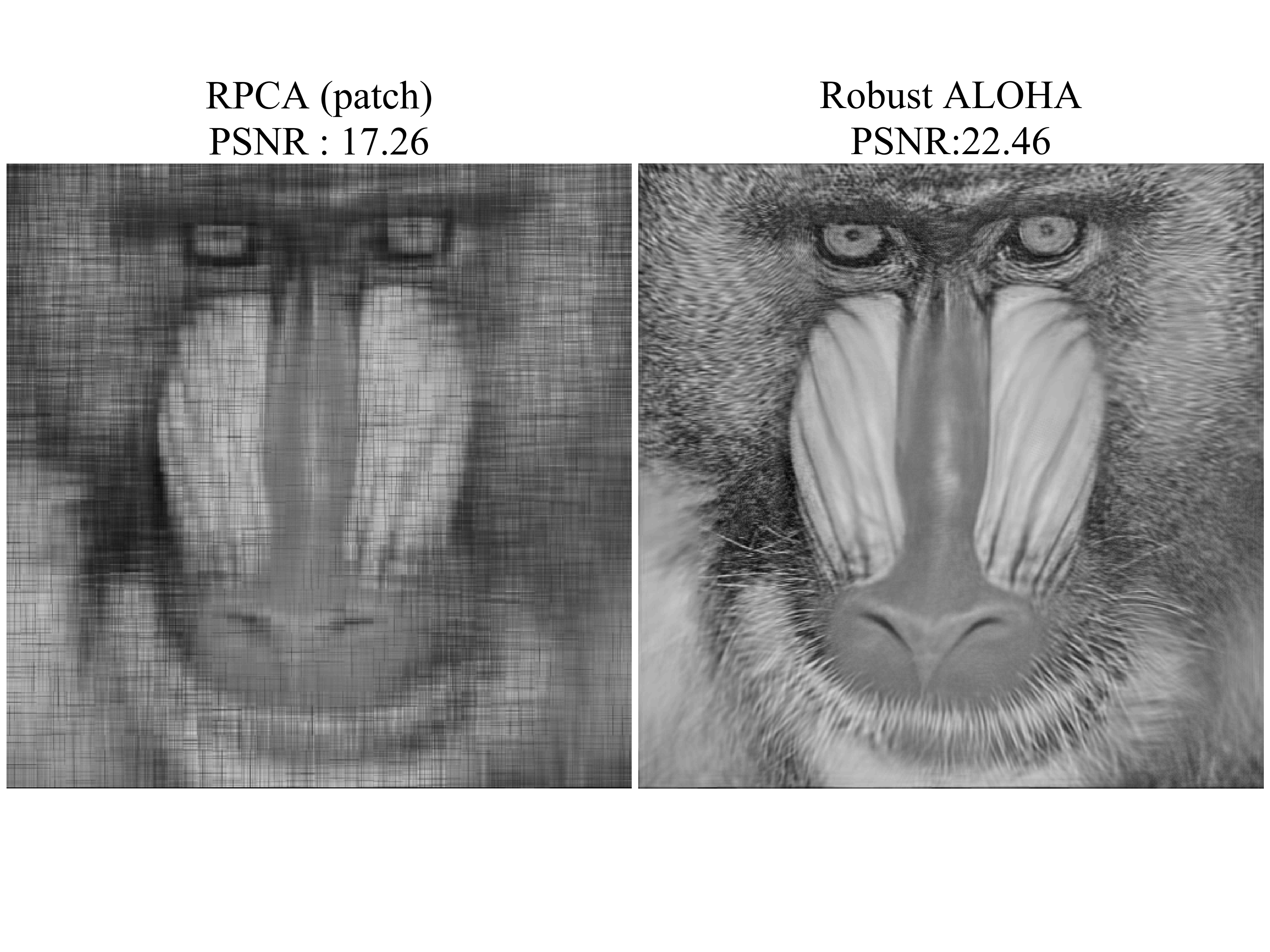}
\caption{Comparison with conventional RPCA approach with the proposed method under 40\% random valued impulse noises.}
\label{fig:res_rpca}
\end{figure}

To verify that a lifting to a Hankel matrix is essential for performance improvement,
we applied the standard RPCA as an  impulse noise removal algorithm  and compared the results.
Two types of RPCA were implemented: one using whole images, and the other using image patch of the same size as  our robust ALOHA.
Note that the standard RPCA uses an image or patches  as they are, without reformulating them  into a Hankel structured matrix.
For RPCA, we used the software packages provided by the original authors in \cite{lin2010augmented}.
We chose the parameters for the best PSNR results in each reconstruction.
% the necessity of Hankel matrix, we compared sparse and low-rank decomposition based on image, local patch, and Hankel matrix from local patch, respectively in Fig. \ref{fig:res_rpca}.   and used same distribution of impulse noises.
  As you can see in Fig. \ref{fig:res_rpca}, 
  two forms of RPCA implementations completely failed in removing impulse noises, and the detailed image structures were distorted.
  On the other hand,  the robust ALOHA provided nearly perfect noise removal. Such a remarkable performance improvement was originated from an image modeling using a low rankness of annihilating filter-based  Hankel matrix, which again confirm the robust ALOHA is a superior image model and denoising algorithm for images corrupted with impulse noises. 

\section{Discussion}\label{sec:discussion}

Recall that the proposed robust ALOHA was performed patch by patch  without  considering additional similar patches. This is of great importance
in contrast to other denoising algorithms that use  low-rank approaches \cite{ji2010robust,ji2011robust,dong2013nonlocal,orchard2008efficient}. While
the authors in \cite{ji2010robust,ji2011robust,dong2013nonlocal,orchard2008efficient}  used the patch-based low-rankness, all methods required additional redundancies from, for example, multiple dynamic frames  \cite{ji2010robust,ji2011robust} or groups of similar spectral patches  \cite{dong2013nonlocal,orchard2008efficient}. Even though such additional  redundant information may introduce a low-rankness, those approaches could not properly perform denoising without utilising such additional redundancies. On the other hand, the robust ALOHA exploits the low-rankness originated from intrinsic spectral sparsity of an image patch, so no additional redundancy is necessary. Thus,  it is more flexible and powerful.

As briefly discussed in the introduction, a recent research \cite{ayazoglu2012fast} successfully demonstrated accurate prediction of target locations under occlusion using sparse $+$ low rank decomposition of Hankel structured matrix. However, unlike our robust ALOHA,   one-dimensional trajectories extracted from video sequences  are required as inputs
to construct the Hankel structured matrix, because the algorithm was derived 
 based on the assumption that those trajectories follow the linear time invariant state-space models, which was first suggested in \cite{ding2008receding}. 
On the other hand, the Hankel structured matrix in robust ALOHA is derived from two dimensional patches by exploiting the spectral domain sparsity; thus, the construction of the Hankel matrix is different from  \cite{ayazoglu2012fast}.
  Moreover, we exploit an SVD-free minimization algorithm \cite{signoretto2013svd} instead of an augmented Lagrangian method (ALM) in \cite{ayazoglu2012fast,ding2008receding} for saving computational burdens. Therefore, we believe that there exist significant differences between the two approaches. 
  
From the sampling theory point of view, a recent paper  by Chen and Chi \cite{chen2014robust} provides a theoretical estimate of fundamental performance  of our robust ALOHA.
%required samples for structured matrix completion with sparse outlier.
 In \cite{chen2014robust}, the authors showed that  the required number of samples, $m$, for the recovery of signal $\hank\{\Xb\}$ with $\Xb \in\Rd^{M\times N}$ corrupted with sparse outlier $\Eb$ 
 using (P) in \eqref{eq:robustEMAC}
 is given by
\begin{equation}\label{eq:sample}
m>c_1 \mu_1^2 c_s^2 r^2 \log^3(MN),
\end{equation}
when the regularization parameter in \eqref{eq:robustEMAC} is given by
$\tau=1/\sqrt{m\log(MN)}$ and the a noise corruption fraction  is smaller than $20\%$.
In \eqref{eq:sample},
 $c_1>0$ is a numerical constant depending on the corruption fraction, $\mu_1$ is incoherence parameter, $c_s:=\max \left\{ \frac{MN}{pq},\frac{MN}{(M-p+1)(N-q+1)} \right\}$ , and $r$ is the rank of single channel Hankel matrix. 
 Because $m=MN$ in our robust ALOHA for impulse noise removal,  the theoretical result  in \eqref{eq:sample}
strongly supports our finding that as long as the spectrum of a patch is sufficiently sparse,  the proposed sparse and low-rank method can
restore the corrupted signal even with significant impulse noises (in our simulation, up to 40 $\%$).

\section{Conclusion}\label{sec:conclusion}

In this paper, we proposed a sparse + low rank decomposition of annihilating filter-based Hankel matrices for impulse noise removal. The new algorithm, called robust ALOHA, 
extends  the conventional RPCA approaches by exploiting the spectral domain sparsity and the associated rank deficient Hankel matrix.
The robust ALOHA was  implemented  using ADMM iteration with initialization  using LMaFit algorithms. 
In our ADMM formulation,   factorization based nuclear norm minimization, was used instead of SVD so that the computational gain is achieved.
We demonstrated that the robust ALOHA outperformed the existing impulse noise removal algorithms up to 8dB. Furthermore, we showed that the robust ALOHA can be used for salt and pepper noises by incorporating the estimated noise locations. In addition, the extension to impulse noise removals from colour channels was very straightforward by concatenating the Hankel structure matrix side by side and imposing the low-rankness.

The superior performance of the robust ALOHA as well as ALOHA inpainting \cite{7127011} clearly shows that  image modeling using annihilating filter based Hankel matrix is a very powerful tool
with many image processing applications.

\section*{Acknowledgment}

The authors would like to thank Dr. Nikolova for sharing source codes of AM-EPR.
This work was supported by Korea Science and Engineering Foundation under Grant NRF-2014R1A2A1A11052491. %NRF-2009-0081089 and NRF-2013M3A9B2076548.

\ifCLASSOPTIONcaptionsoff
  \newpage
\fi



\bibliographystyle{IEEEtran}
\clearpage
\bibliography{totalbiblio_bispl_kh}
\clearpage

\clearpage

\end{document}